\documentclass[runningheads]{llncs}
\usepackage{calc}
\usepackage{graphicx}
\usepackage{csquotes}
\usepackage[dvipsnames]{xcolor}
\usepackage{epsfig}
\usepackage{etoolbox}
\usepackage{mathtools}
\usepackage{amsmath,amssymb}
\usepackage[inline]{enumitem}

\newtoggle{blind}
\newtoggle{nographics}
\newtoggle{nocomments}

\togglefalse{blind}
\togglefalse{nographics}
\togglefalse{nocomments}

\usepackage{tikz}
\usetikzlibrary{calc}
\usetikzlibrary{backgrounds}

\usepackage{array,colortbl,multirow,multicol,booktabs,ctable}
\usepackage{bm,bbm}
\usepackage{subfig}
\usepackage{grffile}
\usepackage{dblfloatfix}
\usepackage{placeins}
\usepackage{letltxmacro}
\usepackage{xifthen}
\usepackage{xparse}
\usepackage[pagebackref=true,breaklinks=true,letterpaper=true,colorlinks,bookmarks=false]{hyperref}
\usepackage{xspace}
\usepackage{float}

\begin{document}
    \pagestyle{headings}
    \mainmatter
    \title{Deep Vectorization of Technical Drawings}
\def\ECCVSubNumber{1978}

%! Suppress = NonMatchingIf
\iftoggle{blind}{
\titlerunning{ECCV-20 submission ID \ECCVSubNumber}
\authorrunning{ECCV-20 submission ID \ECCVSubNumber}
\author{Anonymous ECCV submission}
\institute{Paper ID \ECCVSubNumber}
}{
\titlerunning{Deep Vectorization of Technical Drawings}
\author{Vage Egiazarian\inst{1\thanks{Equal contribution}}\and
Oleg Voynov\inst{1\footnotemark[1]} \and
Alexey Artemov\inst{1} \and
Denis Volkhonskiy\inst{1} \and
Aleksandr Safin\inst{1} \and
Maria Taktasheva\inst{1} \and
Denis Zorin\inst{2,1} \and
Evgeny Burnaev\inst{1}}
\authorrunning{V. Egiazarian and O. Voynov et al.}
\institute{Skolkovo Institute of Science and Technology, 3 Nobel Street, Skolkovo 143026, Russian Federation \and
New York University, 70 Washington Square South, New York NY 10012, USA\\
\email{\{vage.egiazarian, oleg.voinov, a.artemov, denis.volkhonskiy, aleksandr.safin, maria.taktasheva\}@skoltech.ru,
dzorin@cs.nyu.edu, e.burnaev@skoltech.ru}\\
\href{http://adase.group/3ddl/projects/vectorization}{adase.group/3ddl/projects/vectorization}
}
}

    %%% Graphics
\DeclareGraphicsExtensions{.eps,.pdf,.png,.jpg}
%! Suppress = NonMatchingIf
\iftoggle{nographics}{
	\graphicspath{{img.nano/}{img.tiny/}{img/}}

	\makeatletter
	\LetLtxMacro{\includegraphics@orig}{\includegraphics}
	\RenewDocumentCommand{\includegraphics}{ s O{} m }{%
		{\setlength{\fboxsep}{0pt}%
		 \colorbox{lightgray}{\phantom{\IfBooleanTF{#1}{\includegraphics@orig*}{\includegraphics@orig}[#2]{#3}}}%
		}%
	}
	\makeatother
}{
	\graphicspath{{img/}}
}

%%% WIP macros
\newcommand{\Kill}[1]{}
%! Suppress = NonMatchingIf
\iftoggle{nocomments}{
	\newcommand{\DZ}[1]{\ignorespaces}
	\newcommand{\EB}[1]{\ignorespaces}
	\newcommand{\OV}[1]{\ignorespaces}
	\newcommand{\LA}[1]{\ignorespaces}
	\newcommand{\VE}[1]{\ignorespaces}
	\newcommand{\DV}[1]{\ignorespaces}
	\newcommand{\todo}[1]{\ignorespaces}
	\newcommand{\todoshort}[1]{\ignorespaces}
}{
	\newcommand{\DZ}[1]{\textcolor{ForestGreen}{DZ: #1}}
	\newcommand{\EB}[1]{\textcolor{magenta}{EB: #1}}
	\newcommand{\OV}[1]{\textcolor{orange}{OV: #1}}
	\newcommand{\LA}[1]{\textcolor{Plum}{AA: #1}}
	\newcommand{\VE}[1]{\textcolor{blue}{VE: #1}}
	\newcommand{\DV}[1]{\textcolor{Sepia}{DV: #1}}
	\newcommand{\todo}[1]{\textcolor{red}{TODO: #1}}
	\newcommand{\todoshort}[1]{\textcolor{red}{#1}}
    \newcommand{\fixme}[1]{\textcolor{blue}{#1}}
}

%%% Formatting
\setlength{\tabcolsep}{5.5pt}

% we're (slightly) altering the document template here
% in fact, we just replace the italized paragraph heading with the bold one
% we could do this without altering the template, simply by textbf-ing the heading of paragraphs
\makeatletter
\renewcommand\paragraph{\@startsection{paragraph}{4}{\z@}%
{-12\p@ \@plus -4\p@ \@minus -4\p@}%
{-0.5em \@plus -0.22em \@minus -0.1em}%
{\normalfont\normalsize\bfseries\boldmath}}
\makeatother

\def\vs.{vs.\spacefactor=\the\sfcode`\v}
\def\etc.{etc.\spacefactor=\the\sfcode`\c}

\makeatletter
\DeclareRobustCommand\onedot{\futurelet\@let@token\@onedot}
\def\@onedot{\ifx\@let@token.\else.\null\fi\xspace}

\def\eg{\emph{e.g}\onedot} \def\Eg{\emph{E.g}\onedot}
\def\ie{\emph{i.e}\onedot} \def\Ie{\emph{I.e}\onedot}
\def\wrt{w.r.t\onedot} \def\dof{d.o.f\onedot}
\makeatother

%%% Notations and aliases
\def\vec#1{\bm{#1}}
\def\ten#1{#1}

\def\parens#1{\left(#1\right)}
\def\braces#1{\left\{#1\right\}}
\def\brackets#1{\left[#1\right]}
\def\withcondition#1#2{\left.#1\right|_{#2}}

\DeclarePairedDelimiter\norm{\lVert}{\rVert}
\DeclarePairedDelimiter\abs{\lvert}{\rvert}
\def\indicator#1{\mathbbm{1}\brackets{#1}}

\newcommand{\blank}{{{}\cdot{}}}

%% labels
\def\primlabel{\mathrm{prim}}
\def\pixlabel{\mathrm{pix}}
\def\poslabel{\mathrm{pos}}
\def\sizelabel{\mathrm{size}}
\def\rdnlabel{\mathrm{rdn}}

\def\optimized#1{\mathcolor{blue}{#1}}
\makeatletter
\def\mathcolor#1#{\@mathcolor{#1}}
\def\@mathcolor#1#2#3{%
\protect\leavevmode
\begingroup
\color#1{#2}#3%
\endgroup
}
\makeatother

%% indicies and sizes
\def\primitiveindex{k}
\def\secondprimitiveindex{j}
\def\primitivenumber{{n_{\primlabel}}}
\def\kprim{{\primitiveindex_{\primlabel}}}
\def\jprim{{\secondprimitiveindex_{\primlabel}}}

\def\pixelindex{i}
\def\pixelnumber{{n_{\pixlabel}}}
\def\ipix{{\pixelindex_{\pixlabel}}}

\def\batchsize{b}
\def\resnetdepth{{n_{\mathrm{res}}}}
\def\transformerindex{i}
\def\transformerdepth{{n_{\mathrm{dec}}}}
\def\embeddingdim{{d_{\mathrm{emb}}}}

\def\parameterindex{\alpha}

\def\image{I}
\def\patchimage{\image_p}
\def\featuremap{\ten{X}^{\mathrm{im}}}
\def\patchsize{64}

\def\primparameter{\theta}
\def\allprimitiveparametersets{\ten{\Theta}}
\def\primitiveparameterset{\vec{\primparameter}}
\def\primitiveparametersettarget{\hat{\primitiveparameterset}}
\def\allprimitiveparametersetstarget{\hat{\allprimitiveparametersets}}

\def\imagecoordinatex{x}
\def\imagecoordinatey{y}
\def\primitivewidth{w}

\def\transformer{\text{Transformer}}
\def\transformerembedding{\ten{X}^{\mathrm{pr}}}

\def\confidence{p}
\def\confidencetarget{\hat{\confidence}}
\def\allconfidences{\vec{\confidence}}
\def\allconfidencestarget{\vec{\confidencetarget}}

\def\ltwolossweight{\lambda}

%% primitive optimization
\def\charge{q}
\def\chargedistrib{\vec{\charge}}

\def\chargetotal{\charge_\ipix}
\def\chargetotalsub{\charge_{-\primitiveindex,\ipix}}
\def\chargedistribtotal{\chargedistrib}

\def\chargeraster{\hat{\charge}_\ipix}
\def\chargedistribraster{\hat{\chargedistrib}}

\def\chargedistribredun{\chargedistrib^{\rdnlabel}_{\primitiveindex}}
\def\chargeredun{\charge^{\rdnlabel}_{\primitiveindex,\pixelindex}}
\def\chargeredunpix{\charge^{\rdnlabel}_{\primitiveindex,\ipix}}

\def\allprimitiveparametersetsrefined{\allprimitiveparametersets^{\mathrm{ref}}}
\def\primitiveparametersetpos{\primitiveparameterset^{\poslabel}}
\def\allprimitiveparametersetspos{\allprimitiveparametersets^{\poslabel}}
\def\primitiveparametersetsize{\primitiveparameterset^{\sizelabel}}
\def\allprimitiveparametersetssize{\allprimitiveparametersets^{\sizelabel}}

\def\drpotential#1{\varphi\parens{#1}}
\def\cfar{\lambda_{\mathrm{f}}}

\def\coordinatevectorlen{r}
\def\coordinatevector{\vec{\coordinatevectorlen}}
\def\dprimitivearea#1{d r_{#1}^2}
\def\rclose{R_{\mathrm{c}}}
\def\rfar{R_{\mathrm{f}}}

\def\primitivearea{\Omega}
\def\canvasarea{S}

\def\edr{E}
\def\edrpointprim{\edr_{\primitiveindex}}
\def\edrfull{\edr}
\def\edrpos{\edr^{\poslabel}_{\primitiveindex}}
\def\edrsize{\edr^{\sizelabel}_{\primitiveindex}}
\def\edrredun{\edr^{\rdnlabel}_{\primitiveindex}}

\def\edrprimpix{\edr^{\primlabel,\pixlabel}}
\def\edrprimprim{\edr^{\primlabel,\primlabel}}
\def\edrprim{\edr^{\primlabel}}

\def\edrcoef{c}
\def\edrcoefdistrib{\vec{\edrcoef}}

\def\primitivedirection{\vec{l}}
\def\excessvectorfielddir{\vec{m}}
\def\excessvectorfield{\excessvectorfielddir_{\primitiveindex,\pixelindex}}
\def\excessvectorfieldpix{\excessvectorfielddir_{\primitiveindex,\ipix}}
\def\redunbeta{\beta}

%% postprocessing
\def\mergedistance{\rho_+}
\def\nomergedistance{\rho_-}
\def\mergeangle{\alpha_+}
\def\nomergeangle{\alpha_-}

%% metrics
\def\hausdorff{d_{\mathrm{H}}}
\def\meanmin{d_{\mathrm{M}}}
\def\numberofprimitives{\#P}
    \maketitle

\begin{abstract}
    We present a new method for vectorization of technical line drawings, such as floor plans, architectural drawings, and 2D CAD images.
    Our method includes
    (1) a deep learning-based cleaning stage to eliminate the background and imperfections in the image and fill in missing parts,
    (2) a transformer-based network to estimate vector primitives,
    and (3) optimization procedure to obtain the final primitive configurations.
    We train the networks on synthetic data, renderings of vector line drawings, and manually vectorized scans of line drawings.
    Our method quantitatively and qualitatively outperforms a number of existing techniques on a collection of representative technical drawings.
    \keywords{transformer network, vectorization, floor plans, technical drawings}
\end{abstract}

\begin{figure*}[t]
    \centerline{\includegraphics[width=0.9\linewidth]{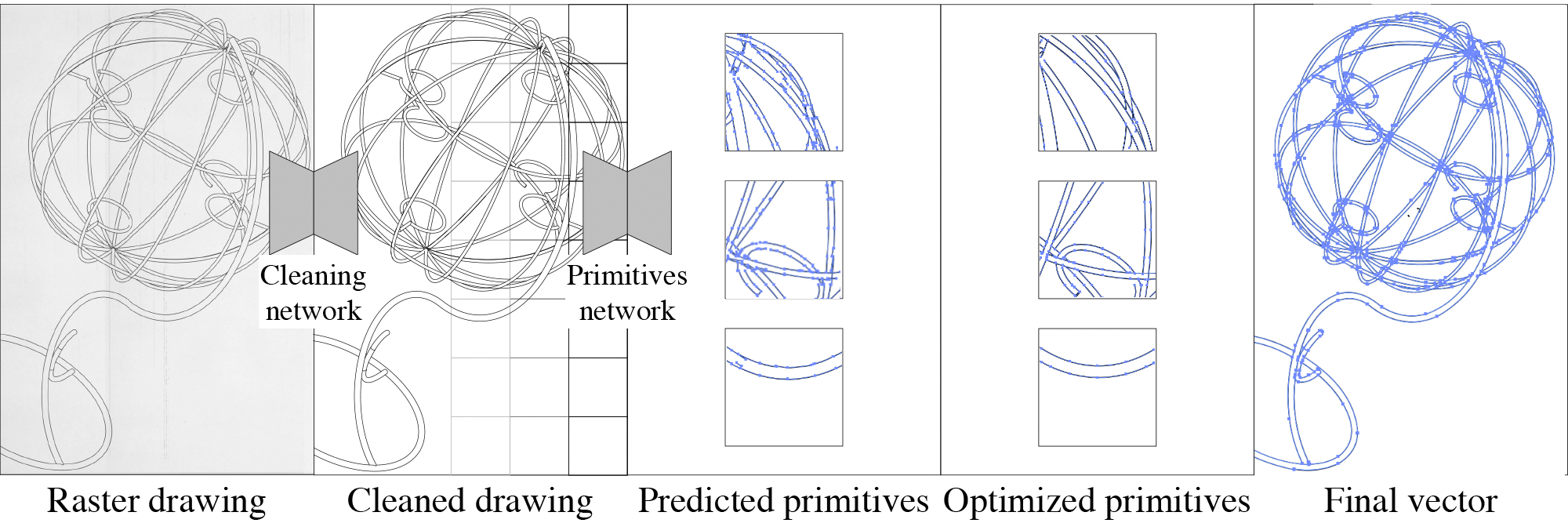}}
    \caption{An overview of our vectorization method.
    First, the input image is cleaned with a deep CNN.
    Then, the clean result is split into patches,
    and primitive placement in each patch is estimated with with a deep neural network.
    After that, the primitives in each patch are refined via iterative optimization.
    Finally, the patches are merged together into a single vector image.}
    \label{fig:framework-overview-eccv}
\end{figure*}

\section{Introduction}
\label{sec:intro}

Vector representations are often used for technical images, such as architectural and construction plans and engineering drawings.  Compared to raster images, vector representations have a number of advantages. They are scale-independent, much more compact, and, most importantly, support easy primitive-level editing. These representations also provide a basis for higher-level semantic structure in drawings (\eg., with sets of primitives hierarchically grouped into semantic objects).

However, in many cases, technical drawings are available only in raster form. Examples include older drawings done by hand, or for which only the hard copy is available, and the sources were lost, or images in online collections.  When the vector representation of a drawing document is unavailable, it is reconstructed, typically by hand, from scans or photos. Conversion of a raster image to a vector representation is usually referred to as \emph{vectorization}.

While different applications have distinct requirements for vectorized drawings, common goals for vectorization are:
\begin{itemize}
    \item approximate the semantically or perceptually important parts of the input image well; 
    \item remove, to the extent possible, the artifacts or extraneous data in the images, such as missing parts of line segments and noise;
    \item minimize the number of used primitives, producing a compact and easily editable representation. 
\end{itemize}

We note that the first and last requirements are often conflicting. \Eg., in the extreme case, for a clean line drawing, 100\% fidelity can be achieved by \textquote{vectorizing} every pixel with a separate line.

In this paper, we aim for geometrically precise and compact reconstruction of vector representations of technical drawings in a fully automatic way.
Distinctive features of the types of drawings we target include the prevalence of simple shapes (line segments, circular arcs, etc.) and relative lack of irregularities (such as interruptions and multiple strokes approximating a single line) other than imaging flaws.
We develop a system which takes as input a technical drawing
and vectorizes it into a collection of line and curve segments (Figure~\ref{fig:framework-overview-eccv}).
Its elements address vectorization goals listed above.
The central element is a deep-learning accelerated optimization method
that matches geometric primitives to the raster image.
This component addresses the key goal of finding a compact representation of a part of the raster image (a \emph{patch}) with few vector primitives.
It is preceded by a learning-based image preprocessing stage,
that removes background and noise and performs infill of missing parts of the image,
and is followed by a simple heuristic postprocessing stage,
that further reduces the number of primitives by merging the primitives in adjacent patches.

Our paper includes the following contributions:
\begin{enumerate}
    \item We develop a novel vectorization method. It is based on a learnable deep vectorization model and a new primitive optimization approach. We use the model to obtain an initial vector approximation of the image, and the optimization produces the final result.
    \item Based on the proposed vectorization method, we demonstrate a complete vectorization system, including a preprocessing learning-based cleaning step and a postprocessing step aiming to minimize the number of primitives.
    \item We conduct an ablation study of our approach and compare it to several state-of-the-art methods.
\end{enumerate}

\section{Related work}
\label{sec:related}

\paragraph{Vectorization.}
There is a large number of methods for image and line drawing vectorization.
However, these methods solve somewhat different, often imprecisely defined versions of the problem
and target different types of inputs and outputs.
Some methods assume clean inputs and aim to faithfully reproduce all geometry in the input,
while others aim, \eg., to eliminate multiple close lines in sketches.
Our method is focused on producing an accurate representation of input images with mathematical primitives.

One of the widely used methods for image vectorization is Potrace~\cite{selinger2003potrace}.
It requires a clean, black-and-white input and extracts boundary curves of dark regions,
solving a problem different from ours
(\eg., a single line or curve segment is always represented by polygon typically with many sides).
Recent works~\cite{munusamy2018vectordefense,kim2018semantic} use Potrace as a stage in their algorithms.

Another widely used approach is based on curve network extraction and topology cleanup~%
\cite{favreau2016fidelity,bessmeltsev2019vectorization,Najgebauer2019InertiabasedFV,chen2015vectorization,chen2018improved,noris2013topology,hilaire2006robust}.
The method of~\cite{favreau2016fidelity} creates the curve network
with a region-based skeleton initialization followed by morphological thinning.
It allows to manually tune the simplicity of the result trading off its fidelity.
The method of~\cite{bessmeltsev2019vectorization} uses a polyvector field (crossfield)
to guide the orientation of primitives.
It applies a sophisticated set of tuneable heuristics
which are difficult to tune to produce clean vectorizations of technical drawings with a low number of primitives.
The authors of~\cite{Najgebauer2019InertiabasedFV} focus on speeding up sketch vectorization without loss of accuracy
by applying an auxiliary grid and a summed area table.
We compare to~\cite{bessmeltsev2019vectorization} and~\cite{favreau2016fidelity}
which we found to be the best-performing methods in this class.

\paragraph{Neural network-based vectorization.}

To get the optimal result, the methods like~\cite{bessmeltsev2019vectorization,favreau2016fidelity}
require manual tuning of hyper-parameters for each individual input image.
In contrast, the neural network-based approach that we opt for is designed to process large datasets without tuning.

The method of~\cite{liu2017raster} generates vectorized, semantically annotated floor plans from raster images
using neural networks.
At vectorization level, it detects a limited set of axis-aligned junctions and merges them,
which is specific to a subset of floor plans (\eg., does not handle diagonal or curved walls).

In~\cite{ellis2018learning} machine learning is used to extract a higher-level representation from a raster line drawing, specifically a program generating this drawing.
This approach does not aim to capture the geometry of primitives faithfully and is restricted to a class of relatively simple diagrams.

A recent work~\cite{Guo2019DeepLD} focuses on improving the accuracy of topology reconstruction.
It extracts line junctions and the centerline image with a two headed convolutional neural network,
and then reconstructs the topology at junctions with another neural network.

The algorithm of~\cite{gao2019deepspline} has similarities to our method: it uses a neural network-based initialization for a more precise geometry fit for B\'ezier curve segments. Only simple input data (MNIST characters) are considered for line drawing reconstruction. The method was also applied to reconstructing 3D surfaces of revolution from images.

An interesting recent direction is generation of sketches using neural networks that learn a latent model representation for sketch images~\cite{ha2018neural,zhou2018learning,kaiyrbekov2019stroke}. In principle, this approach can be used to approximate input raster images, but the geometric fidelity, in this case, is not adequate for most applications.
In~\cite{zhengstrokenet} an algorithm for generating collections of color strokes approximating an input photo is described.
While this task is related to line drawing vectorization it is more forgiving in terms of geometric accuracy and representation compactness.

We note that many works on vectorization focus on sketches. Although the line between different types of line drawings is blurry, we found that methods focusing exclusively on sketches often produce less desirable results for technical line drawings (\eg.,~\cite{favreau2016fidelity} and~\cite{donati2019complete}).

\paragraph{Vectorization datasets.}
Building a large-scale real-world vectorization dataset is costly and
time-consuming~\cite{li2017deep,simo2018mastering}.
One may start from raster dataset and create a vector ground-truth by tracing the lines manually.
In this case, both location and the style may be difficult to match to the original drawing.
Another way is to start from the vector image and render the raster image from it.
This approach does not necessarily produce realistic raster images,
as degradation suffered by real-world documents are known to be challenging to model~\cite{kanungo2000statistical}.
As a result, existing vectorization-related datasets either lack vector annotation (\eg., CVC-FP~\cite{de2015cvc}, Rent3D~\cite{liu2015rent3d}, SydneyHouse~\cite{chu2016housecraft}, and Raster-to-Vector~\cite{liu2017raster} all provide semantic segmentation masks for raster images but not the vector ground truth) or are synthetic (\eg., SESYD~\cite{delalandre2010generation}, ROBIN~\cite{sharma2017daniel}, and FPLAN-POLY~\cite{rusinol2010relational}).

\paragraph{Image preprocessing.}
Building a complete vectorization system based on our approach
requires the initial preprocessing step that removes imaging artefacts.
Preprocessing tools available in commonly used graphics editors
require manual parameter tuning for each individual image.
For a similar task of conversion of hand-drawn sketches into clean raster line drawings
the authors of~\cite{simo2018mastering,sasaki2018learning} use convolutional neural networks trained on synthetic data.
The authors of~\cite{li2017deep} use a neural network to extract structural lines
(\eg., curves separating image regions) in manga cartoon images.
The general motivation behind the network-based approach is that a convolutional neural network automatically adapts
to different types of images and different parts of the image, without individual parameter tuning.
We build our preprocessing step based on the ideas of~\cite{li2017deep,simo2018mastering}.

\paragraph{Other related work.}
Methods solving other vectorization problems include, \eg.,
\cite{zhao2013image,kansal2015vectorization},
which approximate an image with adaptively refined constant color regions with piecewise-linear boundaries;
\cite{mattyus2017deeproadmapper} which extracts a vector representation of road networks from aerial photographs;
\cite{chai2013recovering} which solves a similar problem and is shown to be applicable to several types of images.
These methods use strong build-in priors on the topology of the curve networks.

\section{Our vectorization system}
\label{sec:pipeline}

Our vectorization system, illustrated in Figure~\ref{fig:framework-overview-eccv},
takes as the input a raster technical drawing cleared of text
and produces a collection of graphical primitives defined by the control points and width,
namely line segments and quadratic B\'ezier curves.
The processing pipeline consists of the following steps:
\begin{enumerate}
    \item We preprocess the input image, removing the noise, adjusting its contrast, and filling in missing parts;
    
    \item We split the cleaned image into patches and for each patch estimate the initial primitive parameters;
    
    \item We refine the estimated primitives aligning them to the cleaned raster;
    
    \item We merge the refined predictions from all patches.
\end{enumerate}

\subsection{Preprocessing of the input raster image}
\label{sec:cleaning}

The goal of the preprocessing step is to convert the raw input data into a raster image with clear line structure
by eliminating noise, infilling missing parts of lines,
and setting all background/non-ink areas to white.
This task can be viewed as semantic image segmentation
in that the pixels are assigned the background or foreground class.
Following the ideas of~\cite{li2017deep,simo2018mastering},
we preprocess the input image with U-net~\cite{ronneberger2015u} architecture,
which is widely used in segmentation tasks.We train our preprocessing network in the image-to-image mode with binary cross-entropy loss.

\subsection{Initial estimation of primitives}
\label{sec:vectran}

To vectorize a clean raster technical drawing, we split it into patches
and for each patch independently estimate the primitives with a feed-forward neural network.
The division into patches increases efficiency, as the patches are processed in parallel,
and robustness of the trained model, as it learns on simple structures.

We encode each patch \(\patchimage \in \brackets{0,1}^{\patchsize\times\patchsize}\)
with a ResNet-based~\cite{he2016deep} feature extractor \(\featuremap = \mathrm{ResNet}\parens{\patchimage}\),
and then decode the feature embeddings of the primitives \(\transformerembedding_{\transformerindex}\)
using a sequence of \(\transformerdepth\) Transformer blocks~\cite{vaswani2017attention}
\begin{equation}
	\transformerembedding_{\transformerindex} = \transformer\parens{\transformerembedding_{\transformerindex - 1}, \featuremap}
	\in \mathbb{R}^{\primitivenumber\times\embeddingdim},
	\qquad
	\transformerindex = 1, \ldots, \transformerdepth.
\end{equation}

Each row of a feature embedding represents one of the \(\primitivenumber\) estimated primitives
with a set of \(\embeddingdim\) hidden parameters.
The use of Transformer architecture allows to vary the number of output primitives per patch.
The maximum number of primitives is set with the size of the \(0^\mathrm{th}\) embedding
\(\transformerembedding_{0}\in \mathbb{R}^{\primitivenumber\times\embeddingdim}\),
initialized with positional encoding, as described in~\cite{vaswani2017attention}.
While the number of primitives in a patch is \textit{a priori} unknown,
more than 97\% of patches in our data contain no more than 10 primitives.
Therefore, we fix the maximum number of primitives  and filter out the excess predictions with an additional stage.
Specifically, we pass the last feature embedding to a fully-connected block,
which extracts the coordinates of the control points, the widths of the primitives
\(\allprimitiveparametersets = \braces{\primitiveparameterset_{\primitiveindex} = \parens{\imagecoordinatex_{\primitiveindex,1}, \imagecoordinatey_{\primitiveindex,1}, \ldots, \primitivewidth_{\primitiveindex}}}_{\primitiveindex=1}^{\primitivenumber}\),
and the confidence values \(\allconfidences \in \brackets{0,1}^{\primitivenumber}\).
The latter indicate that the primitive should be discarded if the value is lower than 0.5.
We detail more on the network in supplementary.

\paragraph{Loss function.}
We train the primitive extraction network with the multi-task loss function
composed of binary cross-entropy of the confidence and a weighted sum of $L_1$ and $L_2$ deviations of the parameters
\begin{gather}
	L\parens{\allconfidences, \allconfidencestarget, \allprimitiveparametersets, \allprimitiveparametersetstarget} =
	\frac{1}{\primitivenumber} \sum_{\primitiveindex = 1}^{\primitivenumber} \parens{L_{\text{cls}}\parens{\confidence_{\primitiveindex}, \confidencetarget_{\primitiveindex}} + L_{\text{loc}} \parens{\primitiveparameterset_{\primitiveindex}, \primitiveparametersettarget_{\primitiveindex}}},\\
	L_{\text{cls}}\parens{\confidence_{\primitiveindex}, \confidencetarget_{\primitiveindex}} =
	- \confidencetarget_{\primitiveindex} \log{\confidence_{\primitiveindex}} -
	\parens{1 - \confidencetarget_{\primitiveindex}}\log{\parens{1 - \confidence_{\primitiveindex}}},\\
	L_{\text{loc}} \parens{\primitiveparameterset_{\primitiveindex}, \primitiveparametersettarget_{\primitiveindex}} =
	\parens{1 - \ltwolossweight}\norm{\primitiveparameterset_{\primitiveindex} - \primitiveparametersettarget_{\primitiveindex}}_1 +
	\ltwolossweight\norm{\primitiveparameterset_{\primitiveindex} - \primitiveparametersettarget_{\primitiveindex}}_2^2.
\end{gather}
The target confidence vector \(\allconfidencestarget\) is all ones, with zeros in the end indicating placeholder primitives,
all target parameters \(\primitiveparametersettarget_{\primitiveindex}\) of which are set to zero.
Since this function is not invariant \wrt~to permutations of the primitives and their control points,
we sort the endpoints in each target primitive and the target primitives by their parameters lexicographically.

\subsection{Refinement of the estimated primitives}
\label{sec:primitive-optimization}

We train our primitive extraction network to minimize the average deviation of the primitives on a large dataset.
However, even with small average deviation, individual estimations may be inaccurate.
The purpose of the refinement step is to correct slight inaccuracies in estimated primitives.

To refine the estimated primitives and align them to the raster image,
we design a functional that depends on the primitive parameters and raster image
and iteratively optimize it \wrt~the primitive parameters
\begin{equation}
    \label{eq:fullpotentialdeclaration}
   \allprimitiveparametersetsrefined = \operatornamewithlimits{argmin}_{\allprimitiveparametersets}
    \edrfull\parens{\allprimitiveparametersets, \patchimage}.
\end{equation}
We use physical intuition of attracting charges spread over the area of the primitives and placed in the filled pixels of the raster image.
To prevent alignment of different primitives to the same region, we model repulsion of the primitives.

We define the optimized functional as the sum of three terms per primitive
\begin{equation}
    \label{eq:fullpotential}
    \edrfull\parens{\allprimitiveparametersetspos, \allprimitiveparametersetssize, \patchimage} = \sum_{\primitiveindex=1}^{\primitivenumber} \edrsize + \edrpos + \edrredun,
\end{equation}
where \(\allprimitiveparametersetspos = \braces{\primitiveparametersetpos_{\primitiveindex}}_{\primitiveindex=1}^{\primitivenumber}\)
are the primitive position parameters,
\(\allprimitiveparametersetssize = \braces{\primitiveparametersetsize_{\primitiveindex}}_{\primitiveindex=1}^{\primitivenumber}\)
are the size parameters, and
\(
\primitiveparameterset_{\primitiveindex} = \parens{
\primitiveparametersetpos_{\primitiveindex},
\primitiveparametersetsize_{\primitiveindex}}
\).

We define the position of a line segment by the coordinates of its midpoint and inclination angle,
and the size by its length and width.
For a curve arc, we define the midpoint at the intersection of the curve
and the bisector of the angle between the segments connecting the middle control point and the endpoints. We use the lengths of these segments,
and the inclination angles of the segments connecting the \textquote{midpoint} with the endpoints.

\paragraph*{Charge interactions.}
We base different parts of our functional  on the energy of interaction of unit point charges  \(\coordinatevector_1\), \(\coordinatevector_2\), defined as a sum of close- and far-range potentials
\begin{equation}
    \label{eq:pointpotential}
    \drpotential{\coordinatevector_1,\coordinatevector_2} = \
            e^{-\frac{\norm{\coordinatevector_1 - \coordinatevector_2}^2}{\rclose^2}}\
          + \cfar e^{-\frac{\norm{\coordinatevector_1 - \coordinatevector_2}^2}{\rfar^2}},
\end{equation}
parameters \(\rclose\), \(\rfar\), \(\cfar\) of which we choose experimentally.
The energy of interaction of the uniform positively charged area of the \(\primitiveindex^{\text{th}}\) primitive
\(\primitivearea_{\primitiveindex}\)
and a grid of point charges \(\chargedistrib = \braces{\charge_{\pixelindex}}_{\pixelindex=1}^{\pixelnumber}\)
at the pixel centers \(\coordinatevector_{\pixelindex}\) is then defined by the following equation,
that we integrate analytically for lines 
\begin{equation}
    \label{eq:primitive-to-grid}
    \edrpointprim\parens{\chargedistrib} = \sum\limits_{\pixelindex=1}^{\pixelnumber}\charge_{\pixelindex} %
    \iint\limits_{\primitivearea_{\primitiveindex}}\drpotential{\coordinatevector,\coordinatevector_{\pixelindex}}\dprimitivearea.
\end{equation}
We approximate it for curves as the sum of integrals over the segments of the polyline flattening this curve.

In our functional we use three different charge grids, encoded as vectors of length \(\pixelnumber\):
\(\chargedistribraster\) represents the raster image with charge magnitudes set to intensities of the pixels,
\(\chargedistrib_{\primitiveindex}\) represents the rendering of the \(\primitiveindex^{\text{th}}\) primitive
with its current values of parameters,
and \(\chargedistribtotal\) represents the rendering of all the primitives in the patch.
The charge grids \(\chargedistrib_{\primitiveindex}\) and \(\chargedistribtotal\) are updated at each iteration.

\paragraph*{Energy terms.}
Below, we denote the componentwise product of vectors with~$\odot$,
and the vector of ones of an appropriate size with \(\vec{1}\).

The first term is responsible for growing the primitive to cover filled pixels
and shrinking it if unfilled pixels are covered, with fixed position of the primitive:
\begin{equation}
    \label{eq:szpotential}
    \edrsize =
    \edrpointprim\parens{ %
    \brackets{\chargedistribtotal - \chargedistribraster}\odot \edrcoefdistrib_{\primitiveindex} %
    + \chargedistrib_{\primitiveindex}\odot\brackets{\vec{1}-\edrcoefdistrib_{\primitiveindex}} %
    }.
\end{equation}
The weighting \(\edrcoef_{\primitiveindex,\pixelindex} \in \braces{0,1}\)
enforces coverage of a continuous raster region following the form and orientation of the primitive.
We set \(\edrcoef_{\primitiveindex,\pixelindex}\) to 1 inside the largest region aligned with the primitive with only shaded pixels of the raster,
as we detail in supplementary.
For example, for a line segment, this region is a rectangle centered at the midpoint of the segment
and aligned with it.

The second term is responsible for alignment of fixed size primitives
\begin{equation}
    \label{eq:pospotential}
    \edrpos = 
    \edrpointprim\parens{\brackets{\chargedistribtotal - \chargedistrib_{\primitiveindex} - \chargedistribraster}\odot\brackets{\vec{1} + 3\edrcoefdistrib_{\primitiveindex}}}.
\end{equation}
The weighting here adjusts this term with respect to the first one,
and subtraction of the rendering of the \(\primitiveindex^{\text{th}}\) primitive from the total rendering of the patch ensures that transversal overlaps are not penalized. 

The last term is responsible for collapse of overlapping \emph{collinear} primitives; for this term, we use $\cfar = 0$:
\begin{gather}
    \label{eq:redunpotential}
        \edrredun =
        \edrpointprim\parens{\chargedistribredun},\,
        \chargeredun = \exp{\parens{-\brackets{\abs{\primitivedirection_{\primitiveindex,\pixelindex}\cdot\excessvectorfield} - 1}^2  \redunbeta}}\norm{\excessvectorfield},
\end{gather}
where \(\primitivedirection_{\primitiveindex,\pixelindex}\) is the direction of the primitive
at its closest point to the \(\pixelindex^{\text{th}}\) pixel,
\(\excessvectorfield = \sum_{\secondprimitiveindex\neq\primitiveindex} \primitivedirection_{\secondprimitiveindex,\pixelindex} \charge_{\secondprimitiveindex,\pixelindex} \)
is the sum of directions of all the other primitives weighted \wrt their \textquote{presence},
and \(\redunbeta = \parens{\cos 15^\circ - 1}^{-2}\) is chosen experimentally.

As our functional is based on many-body interactions,
we can use an approximation well-known in physics --- mean field theory.
This translates into the observation that one can obtain an approximate solution of~\eqref{eq:fullpotentialdeclaration}
by viewing interactions of each primitive with the rest as interactions with a static set of charges,
\ie., viewing each energy term \(\edrpos\), \(\edrsize\), \(\edrredun\)
as depending only on the parameters of the \(\primitiveindex^{\text{th}}\) primitive.
This enables very efficient gradient computation for our functional,
as one needs to differentiate each term \wrt.~a small number of parameters only.
We detail on this heuristic in supplementary.
 
We optimize the functional~\eqref{eq:fullpotential} by Adam.
For faster convergence, every few iterations we join lined up primitives by stretching one and collapsing the rest, and move collapsed primitives into uncovered raster pixels.

\subsection{Merging estimations from all patches}
\label{sec:postprocessing}

To produce the final vectorization, we merge the refined primitives from the whole image
with a straightforward heuristic algorithm.
For lines, we link two primitives if they are close and collinear enough but not almost parallel.
After that, we replace each connected group of linked primitives
with a single least-squares line fit to their endpoints.
Finally, we snap the endpoints of intersecting primitives by cutting down the \textquote{dangling} ends
shorter than a few percent of the total length of the primitive.
For B\'ezier curves, for each pair of close primitives we estimate a replacement curve with least squares
and replace the original pair with the fit if it is close enough.
We repeat this operation for the whole image until no more pairs allow a close fit.
We detail on this process in supplementary.

\section{Experimental evaluation}
\label{sec:evaluation}

\begin{figure}[t]
    \centerline{\includegraphics[width=.4\linewidth]{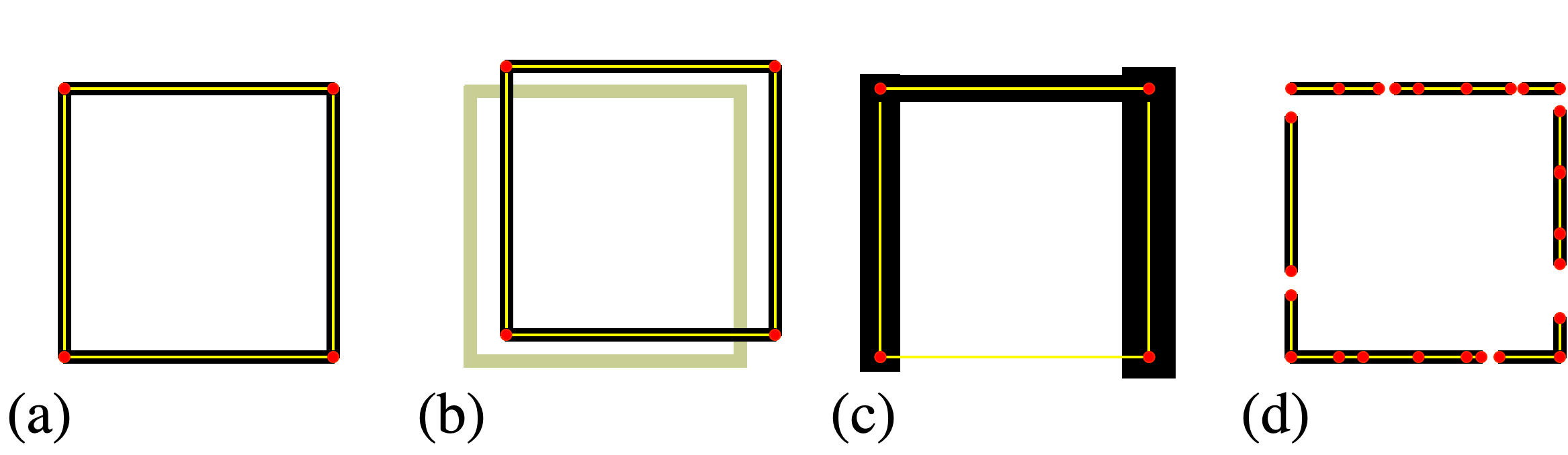}}
    \caption{(a) Ground-truth vector image,
    and artefacts \wrt.~which we evaluate the vectorization performance
    (b) skeleton structure deviation,
    (c) shape deviation,
    (d) overparameterization.}
    \label{fig:failure-mode-capturing}
\end{figure}

We evaluate two versions of our vectorization method:
one operating with lines and the other operating with quadratic B\'ezier curves.
We compare our method against FvS~\cite{favreau2016fidelity},
CHD~\cite{donati2019complete}, and
PVF~\cite{bessmeltsev2019vectorization}.
We evaluate the vectorization performance with four metrics
that capture artefacts illustrated in Figure~\ref{fig:failure-mode-capturing}.

\textbf{Intersection-over-Union (IoU)} reflects deviations in two raster shapes or rasterized vector drawings
$R_1$ and $R_2$ via $\text{IoU}(R_1, R_2) = \frac {R_1 \cap R_2} {R_1 \cup R_2}$.
It does not capture deviations in graphical primitives that have similar shapes but are slightly offset from each other.

\begingroup
\def\vgt{X}
\def\vpred{Y}
\def\dxy{d(x,y)}

\textbf{Hausdorff distance}
\begin{equation}
    \label{eq:hausdorff}
    \small
    \hausdorff\parens{\vgt,\vpred} = \max\braces{\
        \sup\limits_{\tiny x\in\vgt}\inf\limits_{\tiny y\in\vpred}\dxy,\
        \sup\limits_{\tiny y\in\vpred}\inf\limits_{\tiny x\in\vgt}\dxy\
    },
\end{equation}
and \textbf{Mean Minimal Deviation}
\begin{subequations}
    \label{eq:meanmin}
    \small
    \begin{gather}
        \meanmin\parens{\vgt,\vpred} = \frac{1}{2}\parens{\widetilde{d_{\mathrm{M}}}\parens{\vgt\to\vpred} + \widetilde{d_{\mathrm{M}}}\parens{\vpred\to\vgt} }, \\
        \widetilde{d_{\mathrm{M}}}\parens{\vgt\to\vpred} = \left. \int\limits_{\tiny x\in\vgt}\inf\limits_{\tiny y\in\vpred}\dxy d\vgt \middle/ \
            \int\limits_{\tiny x\in\vgt}d\vgt \right.
    \end{gather}
\end{subequations}
measure the difference in skeleton structures of two vector images \(\vgt\) and \(\vpred\),
where \(\dxy\) is Euclidean distance between a pair of points \(x, y\) on skeletons.
In practice, we densely sample the skeletons and approximate these metrics on a pair of point clouds.
\endgroup

\textbf{Number of Primitives}~\numberofprimitives\ measures the complexity of the vector drawing.

\subsection{Clean line drawings}
\label{sec:comparison-pfp-abc}
To evaluate our vectorization system on clean raster images with precisely known vector ground-truth we collected two datasets.

To demonstrate the performance of our method with lines,
we compiled \textbf{PFP vector floor plan dataset}
of 1554 real-world architectural floor plans from a commercial website~\cite{pfpwebsite}.

To demonstrate the performance of our method with curves, we compiled \textbf{ABC vector mechanical parts dataset}
using 3D parametric CAD models from ABC dataset~\cite{koch2019abc}.
They have been designed to model mechanical parts with sharp edges and well defined surface.
We prepared \(\approx 10k\) vector images via projection of the boundary representation of CAD models
with the open-source software Open Cascade~\cite{Opencascade}.

\begin{figure}[t]
    \centerline{\includegraphics[width=0.4\linewidth]{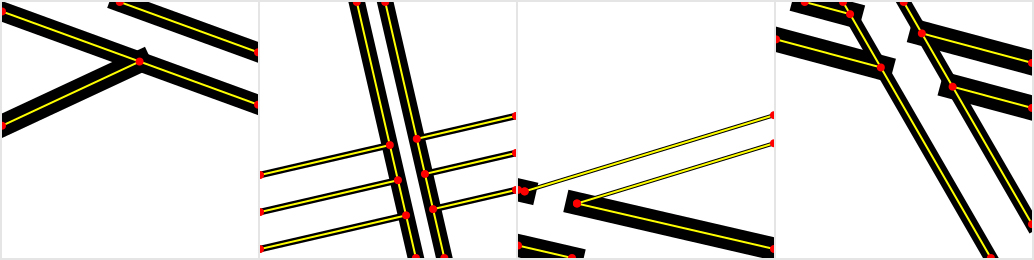}}
    \caption{Examples of synthetic training data for our primitive extraction network.}
    \label{fig:synthetic_local_patches_for_vectran_training}
\end{figure}

We trained our primitive extraction network on random \(64\times 64\) crops, with random rotation and scaling.
We additionally augmented PFP with synthetic data, illustrated in
Figure~\ref{fig:synthetic_local_patches_for_vectran_training}.

For evaluation, we used 40 hold-out images from PFP and 50 images from ABC
with resolution $\sim 2000 \times 3000$ and different complexity per pixel.
We specify image size alongside each qualitative result.
We show the quantitative results of this evaluation in Table~\ref{tbl:vectorization-clean-drawings}
and the qualitative results in Figures~\ref{fig:vectorization-clean-drawings} and~\ref{fig:abc-main}.
Since the methods we compare with produce widthless skeleton,
for fair comparison \wrt.~IoU we set the width of the primitives in their outputs equal to the average on the image.

There is always a trade-off between the number of primitives in the vectorized image and its accuracy,
so the comparison of the results with different number of primitives is not fair.
On PFP, our system outperforms other methods \wrt.~all metrics, and only loses in primitive count to FvS.
On ABC, PVF outperforms our full vectorization system \wrt.~IoU, but not our vectorization method without merging,
as we discuss below in ablation study.
It also produces much more primitives than our method.

\subsection{Degraded line drawings}
\label{sec:comparison-dld}

\begin{figure}[t]
    \centerline{\includegraphics[width=.7\linewidth]{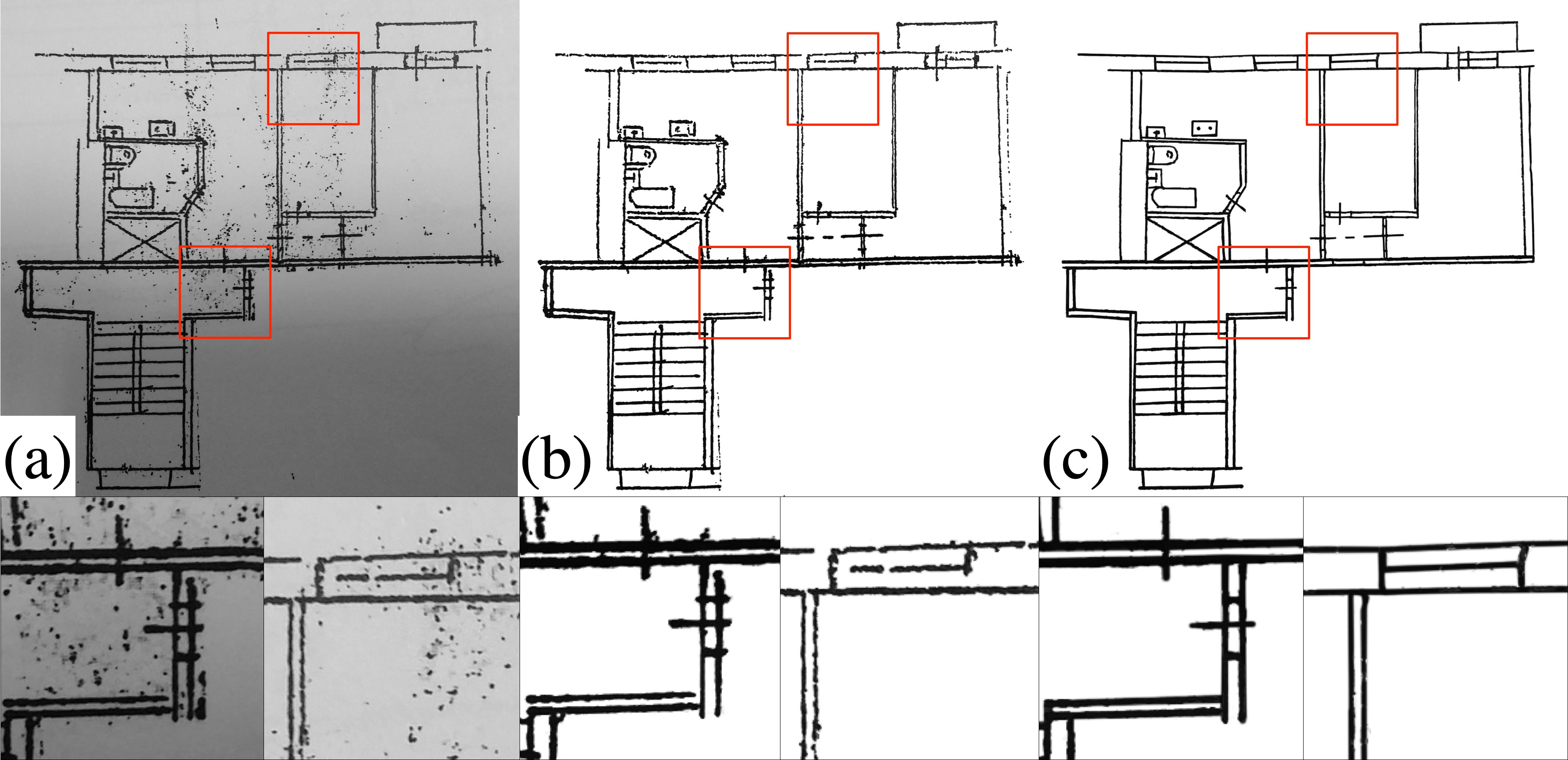}}
    \caption{Sample from DLD dataset: (a) raw input image,
    (b) the image cleaned from background and noise,
    (c) final target with infilled lines.}
    \label{fig:real-cleaning-sample}
\end{figure}

To evaluate our vectorization system on real raster technical drawings,
we compiled \textbf{Degraded line drawings dataset (DLD)}
out of 81 photos and scans of floor plans with resolution \(\sim 1300\times 1000\).
To prepare the raster targets, we manually cleaned each image, removing text, background, and noise,
and refined the line structure, inpainting gaps and sharpening edges
(Figure~\ref{fig:real-cleaning-sample}).

\begingroup
\newcommand\myfigure[2][]{%
\begingroup
\setkeys{myfigure}{#1}%
\centerline{%
\begin{tikzpicture}
    \node[anchor=south west,inner sep=0] (image) at (0,0) {\includegraphics[width=.95\linewidth]{\image}};
    \begin{scope}[x={(image.south east)},y={(image.north west)}]
        \footnotesize
        \def\y{-1.75\baselineskip}
        \def\dx{.1}
        \ifgrid%
        % Guide coordinate grid
        \begingroup
        \fontsize{2}{2}\selectfont
        \draw[help lines,red,xstep=.01,ystep=.01] (0,0) grid (1,1);
        \foreach \x in {0,1,...,99} { \node [anchor=north] at (\x/100,0) {\x}; }
        \foreach \y in {0,1,...,99} { \node [anchor=east] at (0,\y/100) {\y}; }
        \endgroup
        \fi
        % Captions
        \node at (\dx, \y)   [align=center] {FvS~\cite{favreau2016fidelity}          \\ \fvsiou / \fvsdh  \\ \fvsdm / \fvsP};
        \node at (3*\dx, \y) [align=center] {CHD~\cite{donati2019complete}           \\ \chdiou / \chddh  \\ \chddm / \chdP};
        \node at (5*\dx, \y) [align=center] {PVF~\cite{bessmeltsev2019vectorization} \\ \pvfiou / \pvfdh  \\ \pvfdm / \pvfP};
        \node at (7*\dx, \y) [align=center] {Our method                              \\ \ouriou / \ourdh  \\ \ourdm / \ourP};
        \node at (9*\dx, \y) [align=center] {Ground truth, \\ \numberofprimitives\ \gtP};
        % Size
        \begingroup
        \def\armg{.002}
        \fontsize{6}{6}\selectfont
        \draw[latex-latex] (1.005, \splity + \armg) -- (1.005, 1 - \armg);
        \node[rotate=-90] at ($(1.005,0) + (0,\splity)!.5!(0,1) + (.75\baselineskip,0)$) {\fullsize};
        \draw[latex-latex] (1.005, 0 + \armg) -- (1.005, \splity - \armg);
        \node[rotate=-90] at ($(1.005,0) + (0,0)!.5!(0,\splity) + (.75\baselineskip,0)$) {\closesize};
        \endgroup
    \end{scope}
\end{tikzpicture}%
}%
\endgroup
}
\makeatletter
\define@boolkey{myfigure}[]{grid}[false]{}
\define@cmdkey{myfigure}[]{image}{}
\define@cmdkey{myfigure}[]{fvsiou}{}
\define@cmdkey{myfigure}[]{fvsdh}{}
\define@cmdkey{myfigure}[]{fvsdm}{}
\define@cmdkey{myfigure}[]{fvsP}{}
\define@cmdkey{myfigure}[]{chdiou}{}
\define@cmdkey{myfigure}[]{chddh}{}
\define@cmdkey{myfigure}[]{chddm}{}
\define@cmdkey{myfigure}[]{chdP}{}
\define@cmdkey{myfigure}[]{pvfiou}{}
\define@cmdkey{myfigure}[]{pvfdh}{}
\define@cmdkey{myfigure}[]{pvfdm}{}
\define@cmdkey{myfigure}[]{pvfP}{}
\define@cmdkey{myfigure}[]{ouriou}{}
\define@cmdkey{myfigure}[]{ourdh}{}
\define@cmdkey{myfigure}[]{ourdm}{}
\define@cmdkey{myfigure}[]{ourP}{}
\define@cmdkey{myfigure}[]{gtP}{}
\define@cmdkey{myfigure}[]{splity}{}
\define@cmdkey{myfigure}[]{fullsize}{}
\define@cmdkey{myfigure}[]{closesize}{}
\makeatother

\begin{table}[p]
    \setlength\tabcolsep{\widthof{0}*\real{.6}}
    \setlength\aboverulesep{0pt}
    \setlength\belowrulesep{0pt}
    \centering
    \begin{tabular}{l|cccc|cccc|cc}
        \toprule
        \multicolumn{1}{c}{} & \multicolumn{4}{c|}{PFP} & \multicolumn{4}{c|}{ABC} & \multicolumn{2}{c}{DLD}\\
        \midrule
        \multicolumn{1}{c|}{} & \fontsize{6}{7}IoU,\% & $\mathrm{\hausdorff}$, px & $\mathrm{\meanmin}$, px & \numberofprimitives
        &
        \fontsize{6}{7}IoU,\% & $\mathrm{\hausdorff}$, px & $\mathrm{\meanmin}$, px & \numberofprimitives
        &
        \fontsize{6}{7}IoU,\% & \numberofprimitives \\
        \midrule
        FvS~\cite{favreau2016fidelity} & 31 & 381 & 2.8 & 696 &  65 & 38 & 1.7 & 63  &  &  \\
        CHD~\cite{donati2019complete} & 22 & 214 & 2.1 & 1214 & 60 & 9 & 1 & 109  & 47 & 329\\
        PVF~\cite{bessmeltsev2019vectorization} & 60 & 204 & 1.5 & 38k & 89 & 17 & 0.7 & 7818 &  & \\
        \midrule
        Our & 86/88 & 25 & 0.2 & 1331 & 77/77 & 19 & 0.6 & 97  &  79/82 & 452\\
        \bottomrule
    \end{tabular}
    \caption{Quantitative results of vectorization.
    For our method we report two values of IoU: with the average primitive width and with the predicted.}
    \label{tbl:vectorization-clean-drawings}
\end{table}

\begin{figure*}[p]
    \centerline{\begin{tikzpicture}
                    \node[anchor=south west,inner sep=0] (image) at (0,0) {\includegraphics[width=0.95\linewidth]{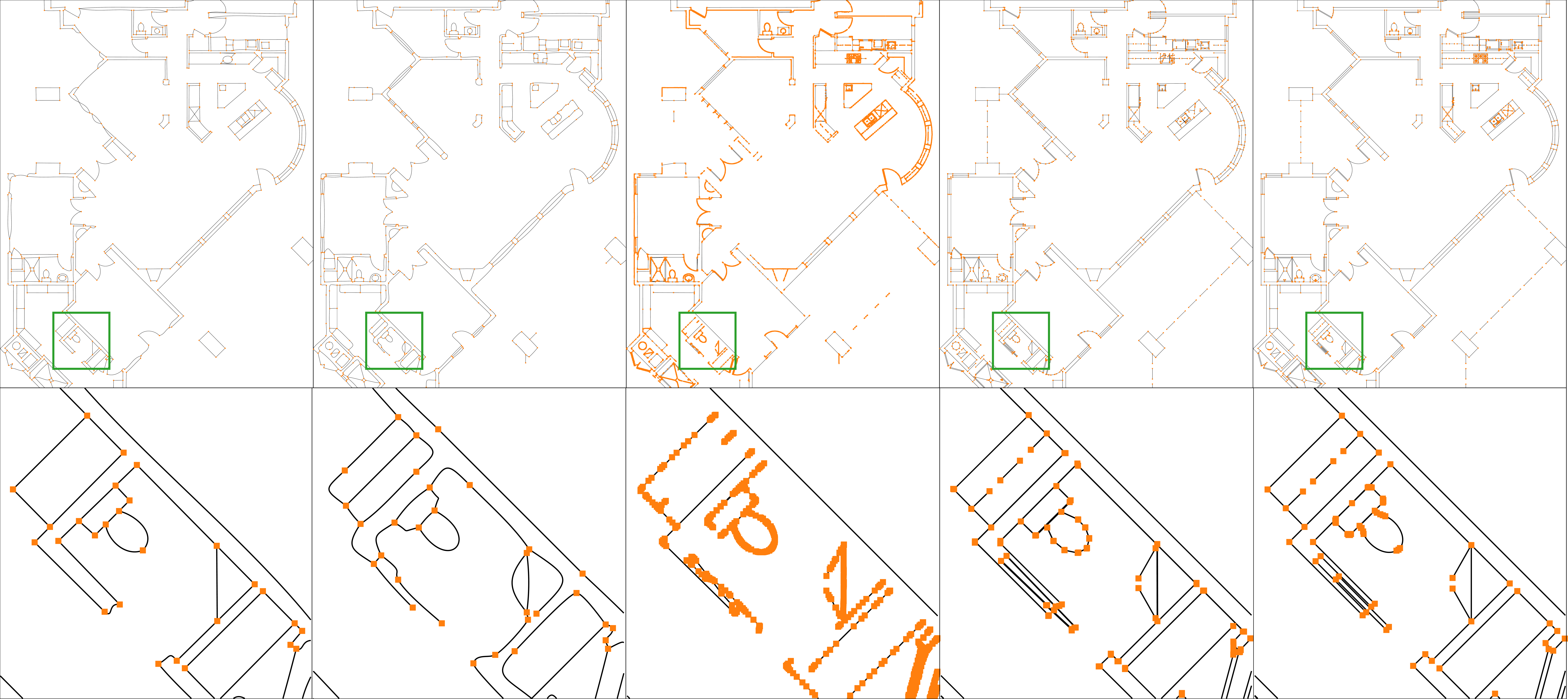}};
                    \begin{scope}[x={(image.south east)},y={(image.north west)}]
                        \footnotesize
                        \def\y{-1.75*\baselineskip}
                        \def\dx{.1}
                        % Captions
                        \node at (\dx, \y)   [align=center] {FvS~\cite{favreau2016fidelity}          \\         29\%  /          415px  \\         4.2px  / \textbf{615} };
                        \node at (3*\dx, \y) [align=center] {CHD~\cite{donati2019complete}           \\         21\%  /          215px  \\         1.9px  / 1192 };
                        \node at (5*\dx, \y) [align=center] {PVF~\cite{bessmeltsev2019vectorization} \\         64\%  /          140px  \\         0.9px  / 35k };
                        \node at (7*\dx, \y) [align=center] {Our method                              \\ \textbf{89\%} / \textbf{  28px} \\ \textbf{0.2px} / 1286 };
                        \node at (9*\dx, \y) [align=center] {Ground truth, \\ \numberofprimitives\ 1634};
                        % Size
                        \begingroup
                        \def\armg{.002}
                        \fontsize{6}{6}\selectfont
                        \draw[latex-latex] (1.005, .445 + \armg) -- (1.005, 1 - \armg);
                        \node[rotate=-90] at ($(1.005, .7225) + (.75\baselineskip,0)$) {1770 px};
                        \draw[latex-latex] (1.005, 0 + \armg) -- (1.005, .445 - \armg);
                        \node[rotate=-90] at ($(1.005, .2225) + (.75\baselineskip,0)$) {256 px};
                        \endgroup
                    \end{scope}
    \end{tikzpicture}}
    \caption{Qualitative comparison on a PFP image,
    and values of IoU / $\mathrm{\hausdorff}$ / $\mathrm{\meanmin}$ / \numberofprimitives\ with best in bold.
    Endpoints of the primitives are shown in orange.}
    \label{fig:vectorization-clean-drawings}
\end{figure*}

\begin{figure*}[p]
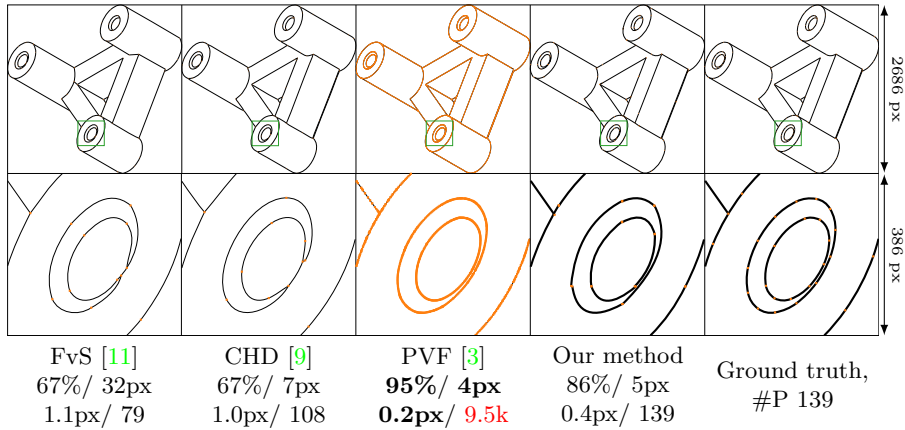

    \myfigure[
    image={abc_1},
    fvsiou={67\%}, fvsdh={32px}, fvsdm={1.1px}, fvsP={79},
    chdiou={67\%}, chddh={7px}, chddm={1.0px}, chdP={108},
    pvfiou={\textbf{95\%}}, pvfdh={\textbf{4px}}, pvfdm={\textbf{0.2px}}, pvfP={\color{red}{9.5k}},
    ouriou={86\%}, ourdh={5px}, ourdm={0.4px}, ourP={139},
    gtP=139,
    splity=.49, fullsize=2686 px, closesize=386 px]{}
    \caption{Qualitative comparison on an ABC image,
    and values of IoU / $\mathrm{\hausdorff}$ / $\mathrm{\meanmin}$ / \numberofprimitives\ with best in bold.
    Endpoints of the primitives are shown in orange.}
    \label{fig:abc-main}
\end{figure*}

\endgroup

To train our preprocessing network,
we prepared the dataset consisting of 20000 synthetic pairs of images of resolution \(512\times 512\).
We rendered the ground truth in each pair from a random set of graphical primitives,
such as lines, curves, circles, hollow triangles, etc.
We generated the input image via rendering the ground truth
on top of one of 40 realistic photographed and scanned paper backgrounds selected from images available online,
and degrading the rendering with random blur, distortion, noise, etc.
After that, we fine-tuned the preprocessing network on DLD.

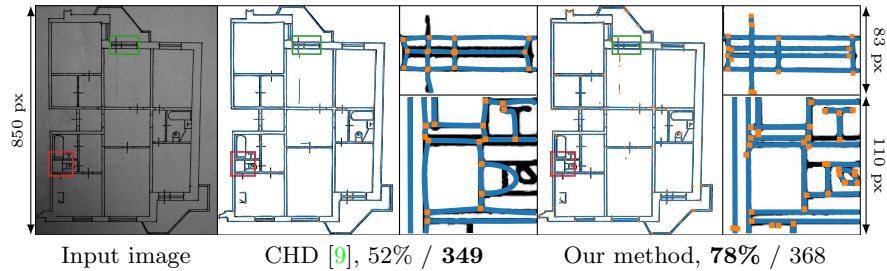
\begin{figure*}[t]
    \centerline{\begin{tikzpicture}
                    \tiny
                    \node[anchor=south west,inner sep=0] (image) at (0,0) {\includegraphics[width=0.9\linewidth]{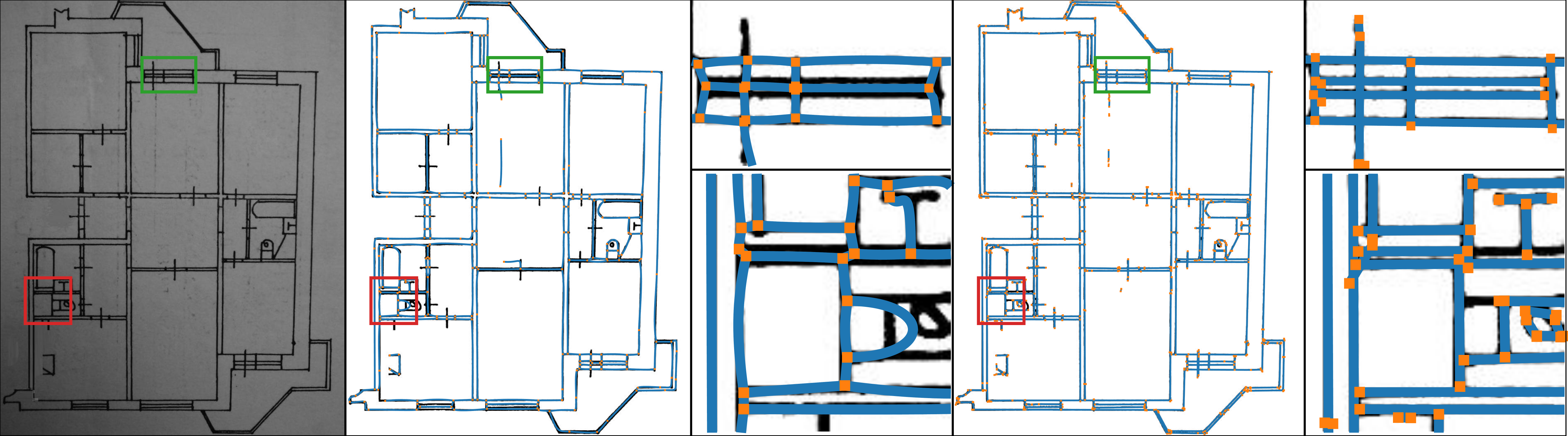}};
                    \begin{scope}[x={(image.south east)},y={(image.north west)}]
                        \footnotesize
                        \def\y{-.75*\baselineskip}
                        \def\inputx{.11}
                        \def\firstx{.41}
                        \def\dx{.3879}
                        % Captions
                        \node at (\inputx, \y) {Input image};
                        \node at (\firstx, \y)     {CHD~\cite{donati2019complete}, 52\%          / \textbf{349}};
                        \node at (\firstx+\dx, \y) {Our method,                    \textbf{78\%} / 368 };
                        % Size
                        \begingroup
                        \def\armg{.002}
                        \fontsize{6}{6}\selectfont
                        \draw[latex-latex] (-.005, 0 + \armg) -- (-.005, 1 - \armg);
                        \node[rotate=90] at ($(-.005, .5) - (.75\baselineskip,0)$) {850 px};
                        \draw[latex-latex] (1.005, .61 + \armg) -- (1.005, 1 - \armg);
                        \node[rotate=-90] at ($(1.005, .805) + (.75\baselineskip,0)$) {83 px};
                        \draw[latex-latex] (1.005, 0 + \armg) -- (1.005, .61 - \armg);
                        \node[rotate=-90] at ($(1.005, .305) + (.75\baselineskip,0)$) {110 px};
                        \endgroup
                    \end{scope}
    \end{tikzpicture}}
    \caption{Qualitative comparison on a real noisy image, and values of IoU / \numberofprimitives\ with best in bold.
    Primitives are shown in blue with endpoints in orange on top of the cleaned raster image.}
    \label{fig:vectorization-noisy-drawings}
\end{figure*}

For evaluation, we used 15 hold-out images from DLD.
We show the quantitative results of this evaluation in Table~\ref{tbl:vectorization-clean-drawings}
and the qualitative results in Figure~\ref{fig:vectorization-noisy-drawings}.
Only CHD allows for degraded input so we compare with this method only.
Since this method produces widthless skeleton,
for fair comparison \wrt.~IoU we set the width of the primitives in its outputs equal to the average on the image,
that we estimate as the sum of all nonzero pixels divided by the length of the predicted primitives.

Our vectorization system outperforms CHD on the real floor plans \wrt.~IoU and produces similar number of primitives.

\begin{figure}[t]
    \centerline{\includegraphics[width=.9\linewidth]{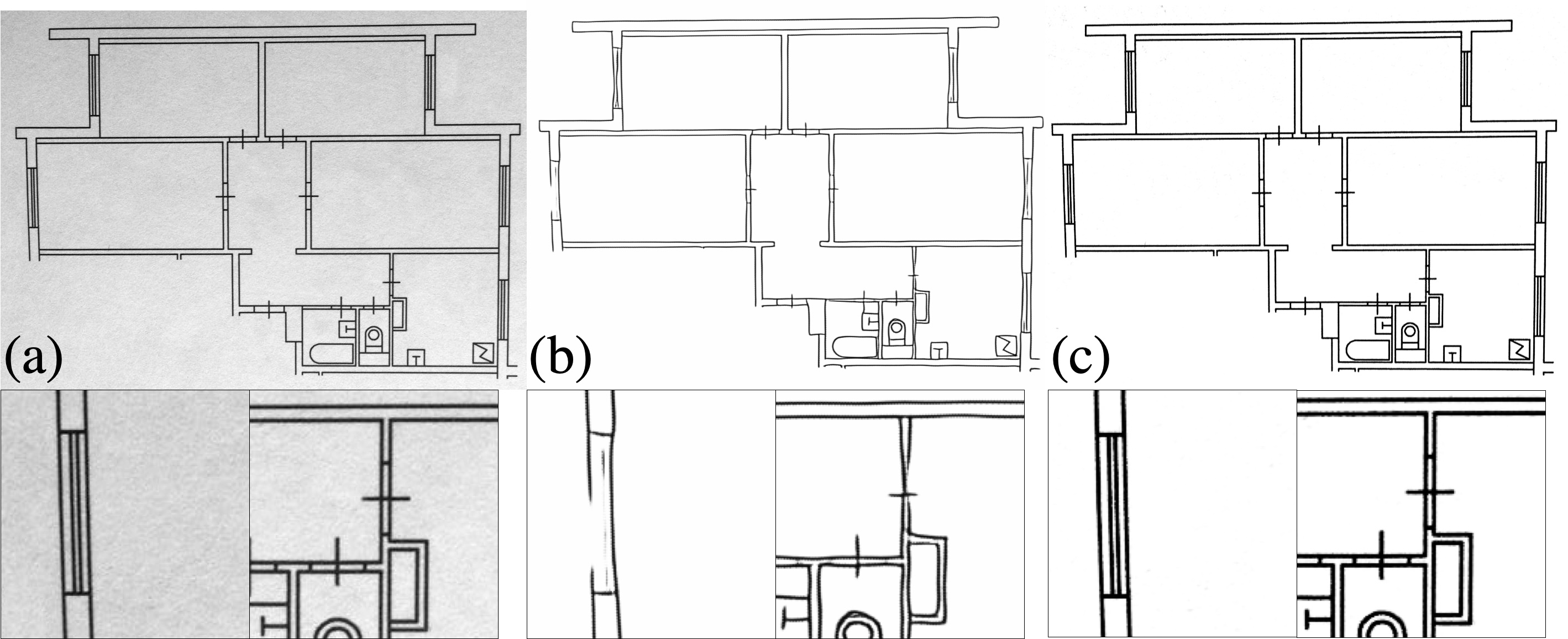}}
    \caption{Example of preprocessing results: (a) raw input image,
    (b) output of MS~\cite{simo2018mastering},
    (c) output of our preprocessing network.
    Note the tendency of MS to combine close parallel lines.}
    \label{fig:cleaning_1}
\end{figure}

\paragraph{Evaluation of preprocessing network.}
We separately evaluate our preprocessing network
comparing with public pre-trained implementation of MS~\cite{simo2018mastering}.
We show the quantitative results of this evaluation in Table~\ref{tbl:cleaning_comparison}
and qualitative results in Figure~\ref{fig:cleaning_1}.
Our preprocessing network keeps straight and repeated lines commonly found in technical drawing
while MS produces wavy strokes and tends to join repeated straight lines, thus harming the structure of the drawing.

\begin{table}[t]
    \setlength\tabcolsep{\widthof{0}*\real{.4}}
    \setlength\aboverulesep{0pt}
    \setlength\belowrulesep{0pt}
    \centering
    \begin{tabular}{l|c|c}
        \toprule
        \multicolumn{1}{c}{} & \fontsize{6}{7}IoU,\% & \fontsize{6}{7} PSNR \\
        \midrule
        MS~\cite{simo2018mastering} & 49 & 15.7 \\
        \midrule
        Our & \textbf{92} & \textbf{25.5} \\
        \bottomrule
    \end{tabular}
    \caption{Quantitative evaluation of the preprocessing step.}
    \label{tbl:cleaning_comparison}
\end{table}

\subsection{Ablation study}
\begingroup

\newcommand\myfigure[2][]{%
\begingroup
\setkeys{myfigure}{#1}%
\centerline{%
\begin{tikzpicture}
    \node[anchor=south west,inner sep=0] (image) at (0,0) {\includegraphics[width=.92\linewidth]{\image}};
    \begin{scope}[x={(image.south east)},y={(image.north west)}]
        \footnotesize
        \def\y{-1.75\baselineskip}
        \def\dx{1./8}
        \ifgrid%
        % Guide coordinate grid
        \begingroup
        \fontsize{2}{2}\selectfont
        \draw[help lines,red,xstep=.01,ystep=.01] (0,0) grid (1,1);
        \foreach \x in {0,1,...,99} { \node [anchor=north] at (\x/100,0) {\x}; }
        \foreach \y in {0,1,...,99} { \node [anchor=east] at (0,\y/100) {\y}; }
        \endgroup
        \fi
        % Captions
        \node at (\dx, \y)   [align=center] {NN               \\ \nniou  / \nndh  \\ \nndm  / \nnP };
        \node at (3*\dx, \y) [align=center] {NN + Refinement  \\ \refiou / \refdh \\ \refdm / \refP };
        \node at (5*\dx, \y) [align=center] {Full             \\ \fuliou / \fuldh \\ \fuldm / \fulP };
        \node at (7*\dx, \y) [align=center] {Ground truth, \\ \numberofprimitives\ \gtP};
        % Size
        \begingroup
        \def\armg{.002}
        \fontsize{6}{6}\selectfont
        \draw[latex-latex] (1.005, \splity + \armg) -- (1.005, 1 - \armg);
        \node[rotate=-90] at ($(1.005,0) + (0,\splity)!.5!(0,1) + (.75\baselineskip,0)$) {\fullsize};
        \endgroup
    \end{scope}
\end{tikzpicture}%
}%
\endgroup
}
\makeatletter
\define@boolkey{myfigure}[]{grid}[false]{}
\define@cmdkey{myfigure}[]{image}{}
\define@cmdkey{myfigure}[]{nniou}{}
\define@cmdkey{myfigure}[]{nndh}{}
\define@cmdkey{myfigure}[]{nndm}{}
\define@cmdkey{myfigure}[]{nnP}{}
\define@cmdkey{myfigure}[]{refiou}{}
\define@cmdkey{myfigure}[]{refdh}{}
\define@cmdkey{myfigure}[]{refdm}{}
\define@cmdkey{myfigure}[]{refP}{}
\define@cmdkey{myfigure}[]{fuliou}{}
\define@cmdkey{myfigure}[]{fuldh}{}
\define@cmdkey{myfigure}[]{fuldm}{}
\define@cmdkey{myfigure}[]{fulP}{}
\define@cmdkey{myfigure}[]{gtP}{}
\define@cmdkey{myfigure}[]{splity}{}
\define@cmdkey{myfigure}[]{fullsize}{}
\makeatother

To assess the impact of individual components of our vectorization system on the results,
we obtained the results on the ABC dataset with the full system,
the system without the postprocessing step,
and the system without the postprocessing and refinement steps.
We show the quantitative results in Table~\ref{tbl:vectorization-ablative-study}
and the qualitative results in Figure~\ref{fig:abl-sup1}.

\begin{table}[t]
    \setlength\tabcolsep{\widthof{0}*\real{.4}}
    \setlength\aboverulesep{0pt}
    \setlength\belowrulesep{0pt}
    \centering
    \begin{tabular}{l|c|c|c|c}
        \toprule
        \multicolumn{1}{c}{} & \fontsize{6}{7}IoU,\% & $\mathrm{\hausdorff}$, px & $\mathrm{\meanmin}$, px & \numberofprimitives \\
        \midrule
        NN  & 65 & 52 & 1.4 & 309 \\
        NN + Refinement & 91 & 19 & 0.3 & 240 \\
        NN + Refinement + Postprocessing & 77 & 19 & 0.6 & 97 \\
        \bottomrule
    \end{tabular}
    \caption{Ablation study on ABC dataset.
    We compare the results of our method with and without refinement and postprocessing}
    \label{tbl:vectorization-ablative-study}
\end{table}

\begin{figure*}
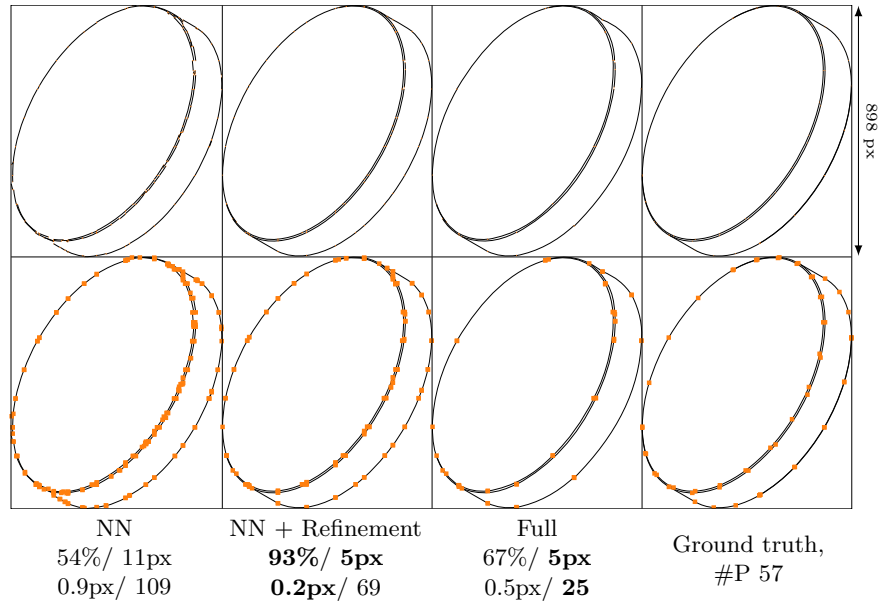

    \myfigure[
    image={supp/abl_1},
    nniou={54\%},           nndh={11px},          nndm={0.9px},           nnP={109},
    refiou={\textbf{93\%}}, refdh={\textbf{5px}}, refdm={\textbf{0.2px}}, refP={69},
    fuliou={67\%},          fuldh={\textbf{5px}}, fuldm={0.5px},          fulP={\textbf{25}},
    gtP=57, splity=.5, fullsize=898 px]{}
    \caption{Results of our method on an ABC image with and without refinement and postprocessing,
    and values of IoU / $\mathrm{\hausdorff}$ / $\mathrm{\meanmin}$ / \numberofprimitives\ with best in bold.
    The endpoints of primitives are shown in orange.}
    \label{fig:abl-sup1}
\end{figure*}

While the primitive extraction network produces correct estimations on average,
some estimations are severely inaccurate, as captured by \(\mathrm{\hausdorff}\).
The refinement step improves all metrics,
and the postprocessing step reduces the number of primitives but deteriorates other metrics
due to the trade-off between number of primitives and accuracy.

We note that our vectorization method without the final merging step outperforms other methods on ABC dataset
in terms of accuracy metrics.

\endgroup

\section{Conclusion}
\label{sec:conclusion}
We presented a four-part system for vectorization of technical line drawings,
which produces a collection of graphical primitives defined by the control points and width.
The first part is the preprocessing neural network that cleans the input image from artefacts.
The second part is the primitive extraction network, trained on a combination of synthetic and real data,
which operates on patches of the image.
It estimates the primitives approximately in the right location most of the time,
however, it is generally geometrically inaccurate.
The third part is iterative optimization, which adjusts the primitive parameters to improve the fit.
The final part is heuristic merging, which combines the primitives from different patches into single vectorized image.
The evaluation shows that our system, in general, performs significantly better compared to a number of recent vectorization algorithms.

Modifications of individual parts of our system would allow it to be applied to different, related tasks.
For example, adjustment of the preprocessing network and the respective training data
would allow for application of our system to extraction of wireframe from a photo.
Modification of the optimized functional and use of the proper training data for primitive extraction network
would allow for sketch vectorization.
Integration with an OCR system would allow for separation and enhancement of text annotations.

\paragraph{Acknowledgements:}
We thank Milena Gazdieva and Natalia Soboleva for their valuable contributions
in preparing real-world raster and vector datasets,
as well as Maria Kolos and Alexey Bokhovkin for contributing parts of shared codebase used throughout this project.
We acknowledge the usage of Skoltech CDISE HPC cluster Zhores for obtaining the presented results.
The work was partially supported by Russian Science Foundation under Grant 19-41-04109.
\FloatBarrier

    \bibliographystyle{splncs04}
    \bibliography{src/bib}
    \newpage
    \appendix
% \maketitle

\section*{Appendix}
We provide additional details on the neural network architecture and training process in Sections~\ref{sec:nnarchitecture} and~\ref{sec:traingdet}.
Details of our postprocessing can be found in Section~\ref{sec:sup.merging}.
We compare runtimes of the methods in Section~\ref{sec:sup.time_vs_accuracy}.
In Section~\ref{sec:sup.on_patches} we show the performance of~\cite{bessmeltsev2019vectorization,donati2019complete,favreau2016fidelity} on small patches in comparison to whole images.
We show example results of our system for cartoon drawings in Section~\ref{sec:sup.cartoon}.
We provide additional comparisons from the ablation study in Section~\ref{sec:ablstudy}
and additional comparisons with other methods in Section~\ref{sec:aresults}.
In Section~\ref{sec:refalg} we describe our refinement algorithm in detail.

\section{Neural Networks architectures}
\label{sec:nnarchitecture}

For image cleaning we use U-net~\cite{ronneberger2015u} encoder-decoder architecture.
It consists of blocks of layers, each containing convolutional and batch normalization layers and ReLU activations.
We use seven such blocks interleaved with MaxPool downsampling in the encoder,
and seven blocks interleaved with nearest neighbor upsampling in the decoder.
We connect the blocks of the encoder and decoder with the same resolution of feature maps with skip connections,
as in the original U-net.

\begin{figure*}[h!]
    \centerline{\includegraphics[width=0.6\linewidth]{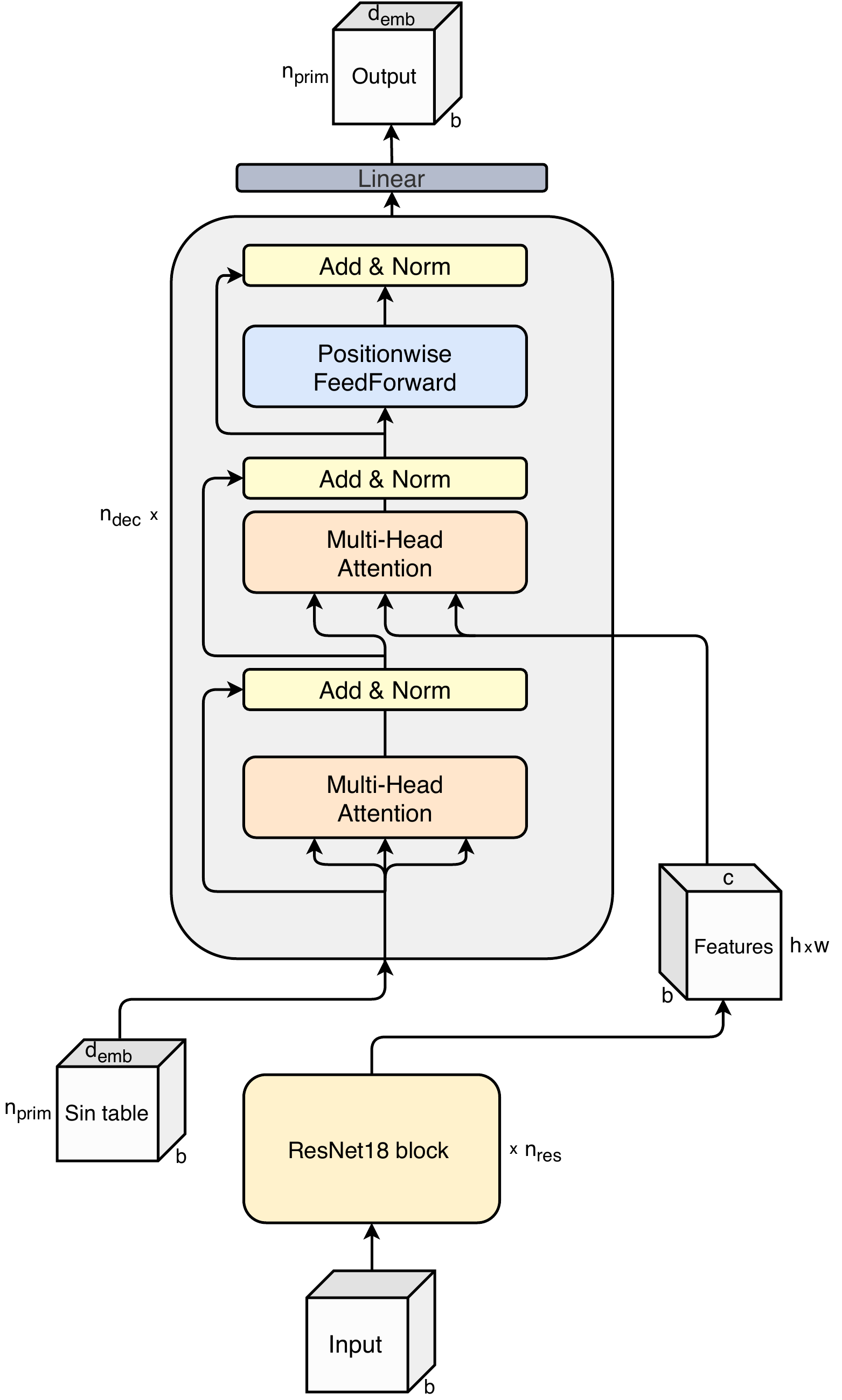}}
    \caption{Architecture of our vectorization network.
    A batch of \(\batchsize\) grayscale raster patches
    is first encoded with the sequence of \(\resnetdepth\) ResNet blocks.
    Then, \(c\) channel feature maps of size \(h\times w\)
    are decoded with a sequence of \(\transformerdepth\) Transformer blocks.
    Finally, the output of the last Transformer block
    is converted with a linear layer into \(\primitivenumber\) sets of primitive parameters per sample in batch.}
    \label{fig:vectorization-architecture}
\end{figure*}

We build our primitive extraction network from two parts:
the encoder consisting of ResNet18 blocks~\cite{He_2016_CVPR},
which extracts features from the raster,
and the decoder Transformer model~\cite{vaswani2017attention},
which estimates the primitive parameters.
The architecture of our primitive extraction network is shown in Figure~\ref{fig:vectorization-architecture}.

We use \(\resnetdepth = 1\) ResNet18 block with \(c = 64\) channels in each convolution.
We use \(\transformerdepth = 8\) Transformer blocks
with \(4\) heads of multi-head attention and \(512\) neurons in the last fully-connected layer.

We set the hidden dimensionality of the primitive representations in the Transformer part of the network
\(\embeddingdim\) equal to the number of primitive parameters, \(6\) for lines and \(8\) for curves:
one for the width, one for the confidence value and the rest for coordinates of the control points.
We keep the other hyper-parameter values the same for lines and curves.

\section{Training details}
\label{sec:traingdet}
  We used Pytorch 1.2 for GPU computations and model training \cite{NEURIPS2019_9015} 
and GNU Parallel to speed up the metric calculations \cite{Tange2011a} for our methods. We trained our models for $15$ epochs on ABC dataset and $17$ epochs on PFP dataset. The batch size was $128$.
We used Adam \cite{kingma2014adam} for optimization with a scheduler with the same hyperparameters as in the original Transformer paper \cite{vaswani2017attention}. It took us approximately four days to train each model on a single Nvidia v100. To speed up the training, we pre-calculated all data augmentations, including cropping, and trained our model on this augmented data. First, we split original images into train, validation, and test sets. Then we cropped and augmented images to prevent  overfitting. 

\begin{figure*}[h!]
    \centerline{\begin{tikzpicture}
                    \tiny
                    \node[anchor=south west,inner sep=0] (image) at (0,0) {\includegraphics[width=0.87\linewidth]{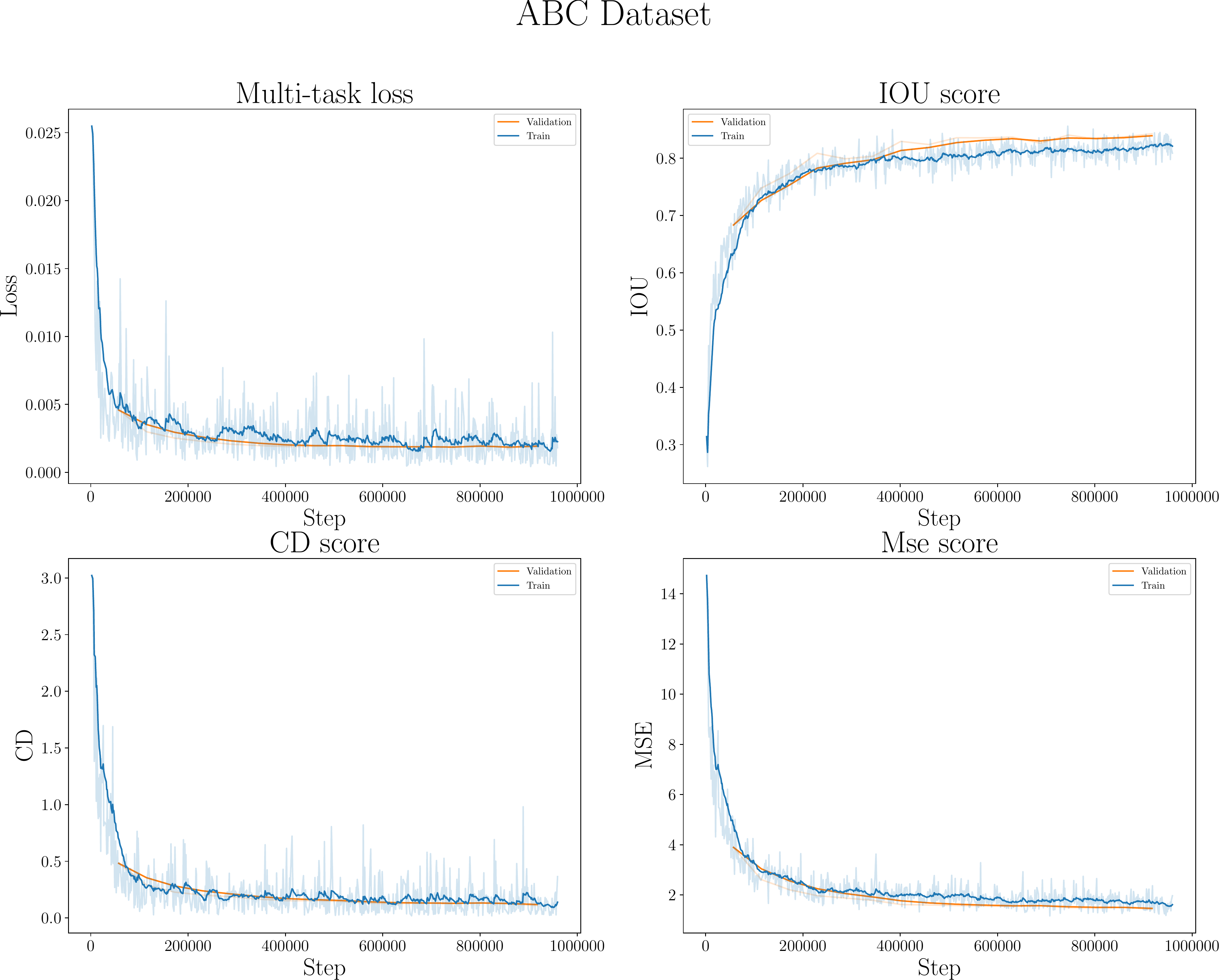}};
    \end{tikzpicture}}
    \caption{Metrics and loss function for train and validation on ABC dataset. One step of X-axis represents calculations on a single batch.}
    \label{fig:abc-graphics}
\end{figure*}

\begin{figure*}[h!]
    \centerline{\begin{tikzpicture}
                    \tiny
                    \node[anchor=south west,inner sep=0] (image) at (0,0) {\includegraphics[width=0.87\linewidth]{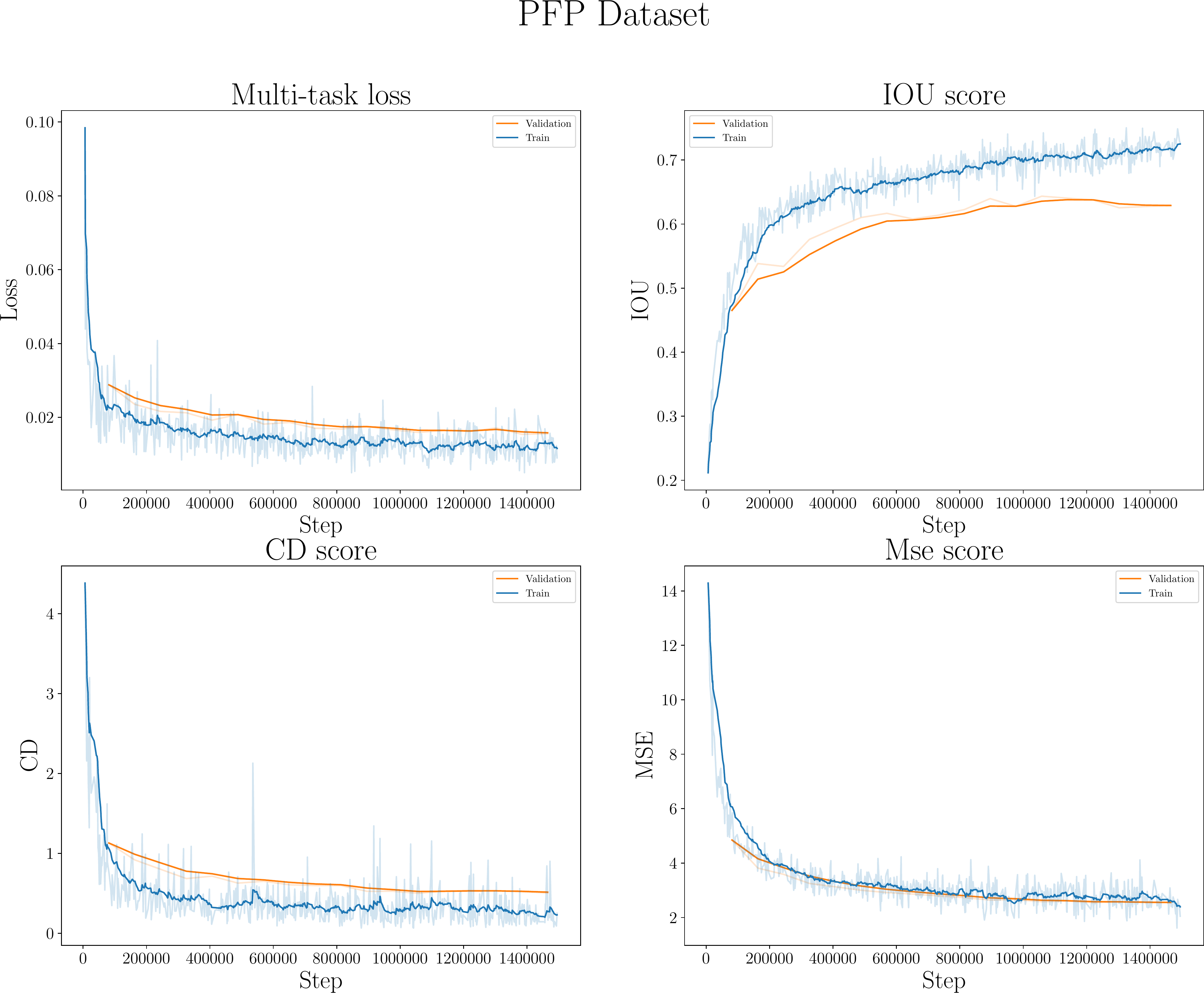}};
    \end{tikzpicture}}
    \caption{Metrics and loss function for train and validation on PFP dataset. One step of the X-axis represents computations on a single batch.}
    \label{fig:pfp-graphics}
\end{figure*}

In Figure~\ref{fig:abc-graphics} and Figure~\ref{fig:pfp-graphics} we provide metrics on train and validation sets for patches with size $64\times 64$ from ABC and PFP datasets correspondingly.

\section{Merging algorithm}
\label{sec:sup.merging}

\begin{figure}[t]
    \centerline{\includegraphics[width=\linewidth]{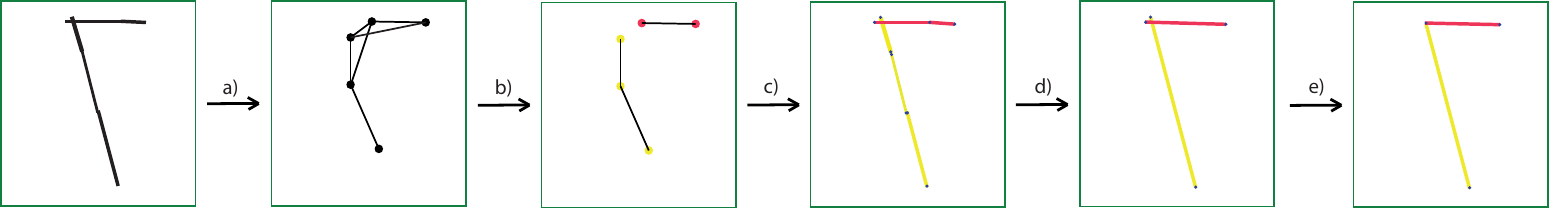}}
    \caption{Our algorithm of line merging:
    (a) we find close lines,
    (b,~c) we join them in the connected components of the graph,
    (d) we fit the endpoints of the lines in each connected component with least squares,
    (e) and finally snap the endpoints of the lines.}
    \label{fig:sup_merge_line}
\end{figure}

For lines we start by building a graph with the primitives as nodes
and edges between nodes that correspond to a pair of lines
that are close and collinear enough but not almost parallel
(Figure~\ref{fig:sup_merge_line}~(a,~b)).
Then, we replace the lines in each connected component of the graph
with a single least-squares line fit to their endpoints
(Figure~\ref{fig:sup_merge_line}~(c,~d)).
Finally, we snap the endpoints of intersecting primitives by cutting down the \textquote{dangling} ends
shorter than a few percent of the total length of the primitive.
(Figure~\ref{fig:sup_merge_line}~(e)).

\begin{figure}[t]
    \centerline{\includegraphics[width=\linewidth]{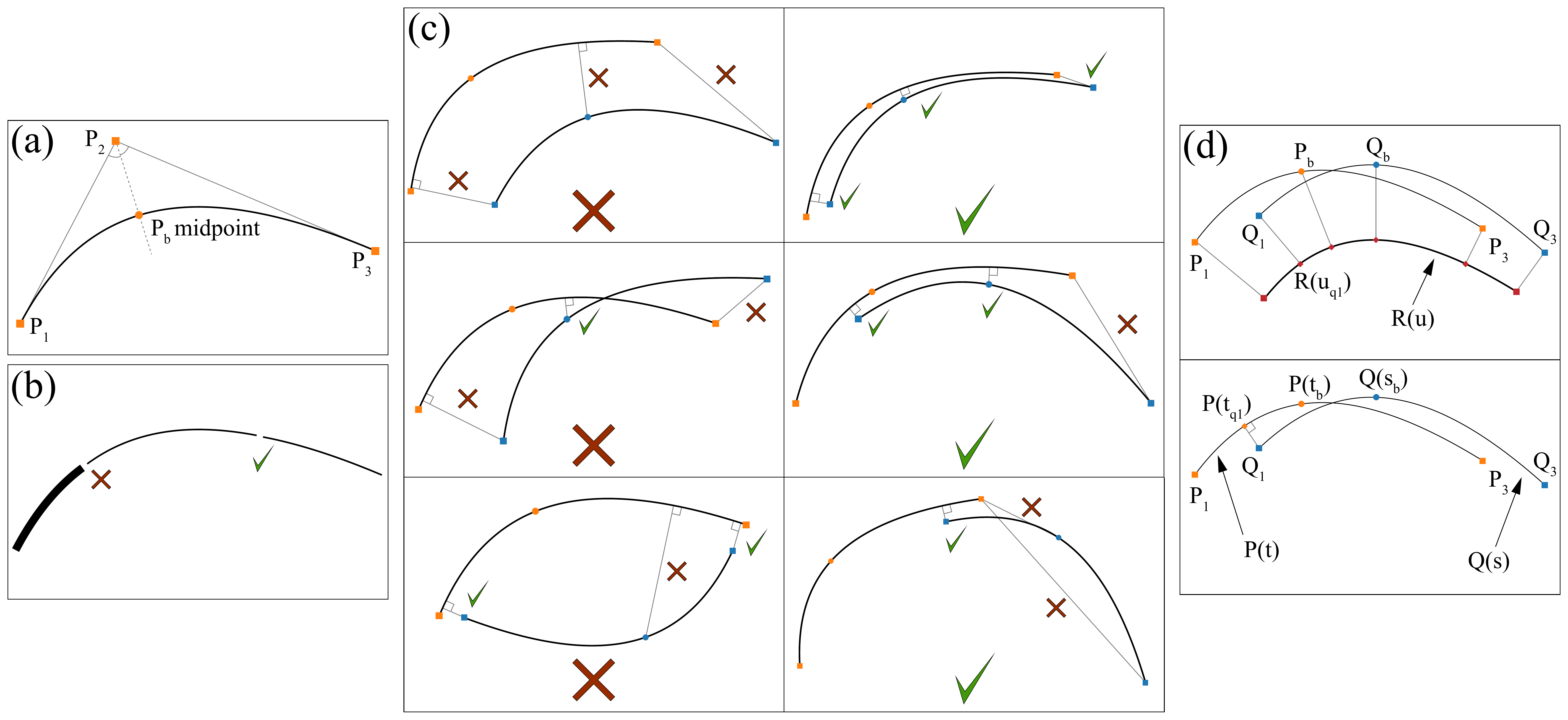}}
    \caption{(a) Our definition of the \textquote{midpoint} for quadratic B\'ezier curve,
    and (b-d) single step of our algorithm of curve merging:
    (b) we check that the widths are close,
    (c) we check that the curves are close,
    (d) we fit the endpoints and midpoints of the curves with least squares.}
    \label{fig:sup_merge_curve}
\end{figure}

For quadratic B\'ezier curves, we iteratively try to replace pairs of curves with a single one.
For each pair of curves
\(\mathrm{P}\parens{t}\), \(t \in \brackets{0,1}\),
\(\mathrm{Q}\parens{s}\), \(s \in \brackets{0,1}\),
we first check if their widths are close
(Figure~\ref{fig:sup_merge_curve}~(b)).
Then, we check if the \textquote{midpoint}
(Figure~\ref{fig:sup_merge_curve}~(a))
and the endpoints of the second curve are close to the first one,
as illustrated in Figure~\ref{fig:sup_merge_curve}~(c).
If all checks are passed, we find a new quadratic B\'ezier curve
\(\mathrm{R}\parens{u}\), \(u \in \brackets{0,1}\)
as a least-squares fit to the endpoints and midpoints of the curves in the pair
(Figure~\ref{fig:sup_merge_curve}~(d)).
Specifically, we minimize the distances between the points
\begin{equation}
    \begin{gathered}
        \mathrm{P_1} = \mathrm{P}\parens{0},\quad
        \mathrm{P_b} = \mathrm{P}\parens{t_\mathrm{b}},\quad
        \mathrm{P_3} = \mathrm{P}\parens{1},\\
        \mathrm{Q_1} = \mathrm{Q}\parens{0},\quad
        \mathrm{Q_b} = \mathrm{Q}\parens{s_\mathrm{b}},\quad
        \mathrm{Q_3} = \mathrm{Q}\parens{1}
    \end{gathered}
\end{equation}
and the points on the new curve
\begin{equation}
    \begin{gathered}
        \mathrm{R}\parens{0},\quad
        \mathrm{R}\parens{t_\mathrm{b} u_\mathrm{q1} / t_\mathrm{q1}},\quad
        \mathrm{R}\parens{u_\mathrm{q1} / t_\mathrm{q1}},\\
        \mathrm{R}\parens{u_\mathrm{q1}},\quad
        \mathrm{R}\parens{1 - \parens{1 - s_\mathrm{b}} \parens{1 - u_\mathrm{q1}}},\quad
        \mathrm{R}\parens{1}
    \end{gathered}
\end{equation}
respectively \wrt.~control points of the new curve.
Here, \(t_\mathrm{b}\) and \(t_\mathrm{q1}\) are the parameter values of
\(\mathrm{P_b}\) and the projection of \(\mathrm{Q_1}\) on the first curve,
\(s_\mathrm{b}\) is the parameter value of \(\mathrm{Q_b}\) on the second curve,
\(u_\mathrm{q1}\) is the parameter value of \(\mathrm{Q_1}\) on the new curve.
We find the value of \(u_\mathrm{q1}\) with brute-force search and take the best fit.
Finally, if the best fit is close enough, we replace the pair of the curves with the fit.
We repeat this process until no more pairs allow for a close fit.

\section{Computation time}
\label{sec:sup.time_vs_accuracy}
Our refinement step is iterative and allows trading longer computation times for more accurate results.
In Table~\ref{tbl:computation_times} we show example computation times for the prior work along with IoU values,
and the computation times required by our system to reach similar IoU values.

Our system without the final merging step reaches the same IoU value as CHD~\cite{donati2019complete} in a similar time,
and the same IoU values as FvS~\cite{favreau2016fidelity} and PVF~\cite{bessmeltsev2019vectorization} in much less time.
We note however that none of the methods were optimized for performance
and that we run the methods in different environment because of technical requirements.

\begin{table}[!ht]
    \setlength\aboverulesep{0pt}
    \setlength\belowrulesep{0pt}
    \centering
    \begin{tabular}{l|c|c|c}
        \toprule
        \multicolumn{1}{c}{} & IoU, \% & Time & \numberofprimitives \\
        \midrule
        CHD~\cite{donati2019complete}           & 64 & 10 s   & 994 \\
        FvS~\cite{favreau2016fidelity}          & 74 & 17.5 m & 433 \\
        PVF~\cite{bessmeltsev2019vectorization} & 91 & 25 h   & 43k \\
        \midrule
        Our, w/o final merging                  & 68 & 35 s   & 2108 \\
        Our, w/o final merging                  & 75 & 50 s   & 2106 \\
        Our, w/o final merging                  & 91 & 5.5 m  & 1502 \\
        \midrule
        Our, w/o final merging, converged       & 92 & 12 m   & 1435 \\
        Our, with final merging                 & 76 & 26 m   & 579  \\
        \bottomrule
    \end{tabular}
    \caption{IoU, computation time, and number of primitives for the results on the Globe (Figure~\ref{fig:abc-sup})
    produced by the prior work,
    for intermediate results of our method with similar values of IoU,
    and for our final result.}
    \label{tbl:computation_times}
\end{table}

\section{Prior work on patches}
\label{sec:sup.on_patches}
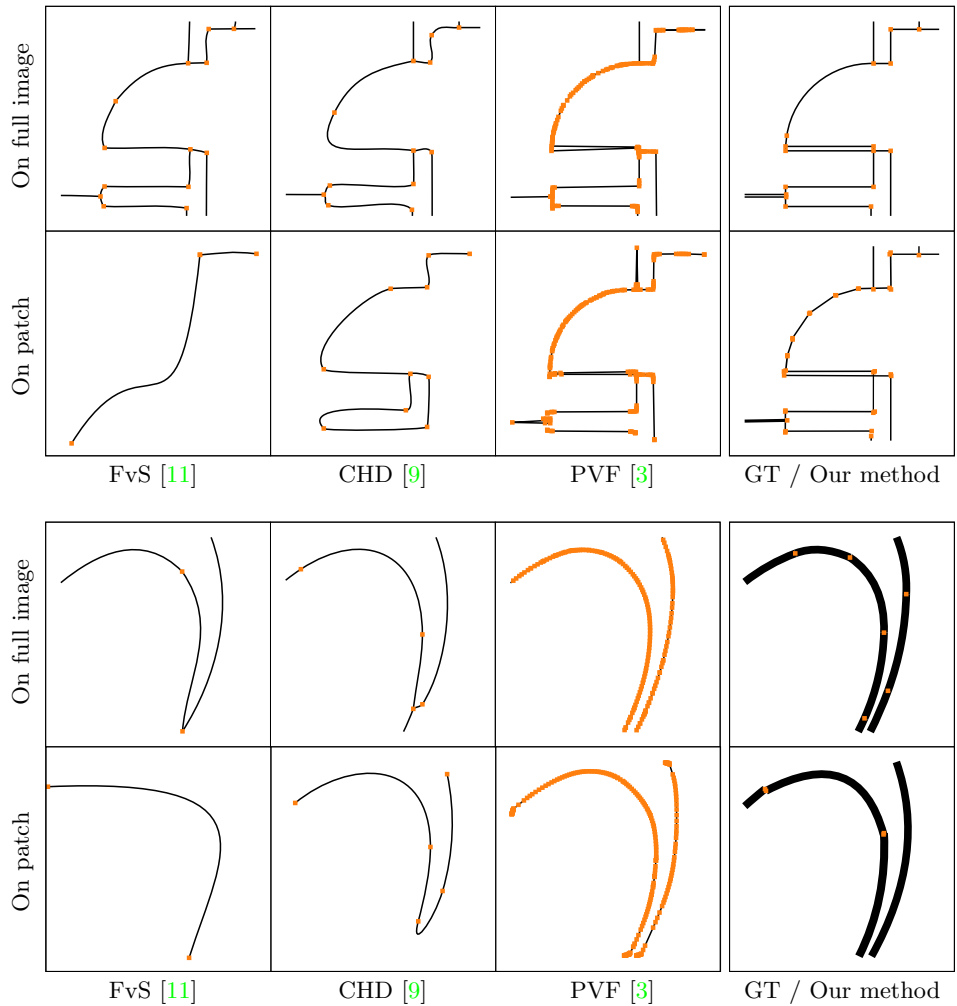
\begin{figure*}[h!]
    \centerline{\begin{tikzpicture}
                    \node[anchor=south west,inner sep=0] (image) at (0,0) {\includegraphics[width=\linewidth]{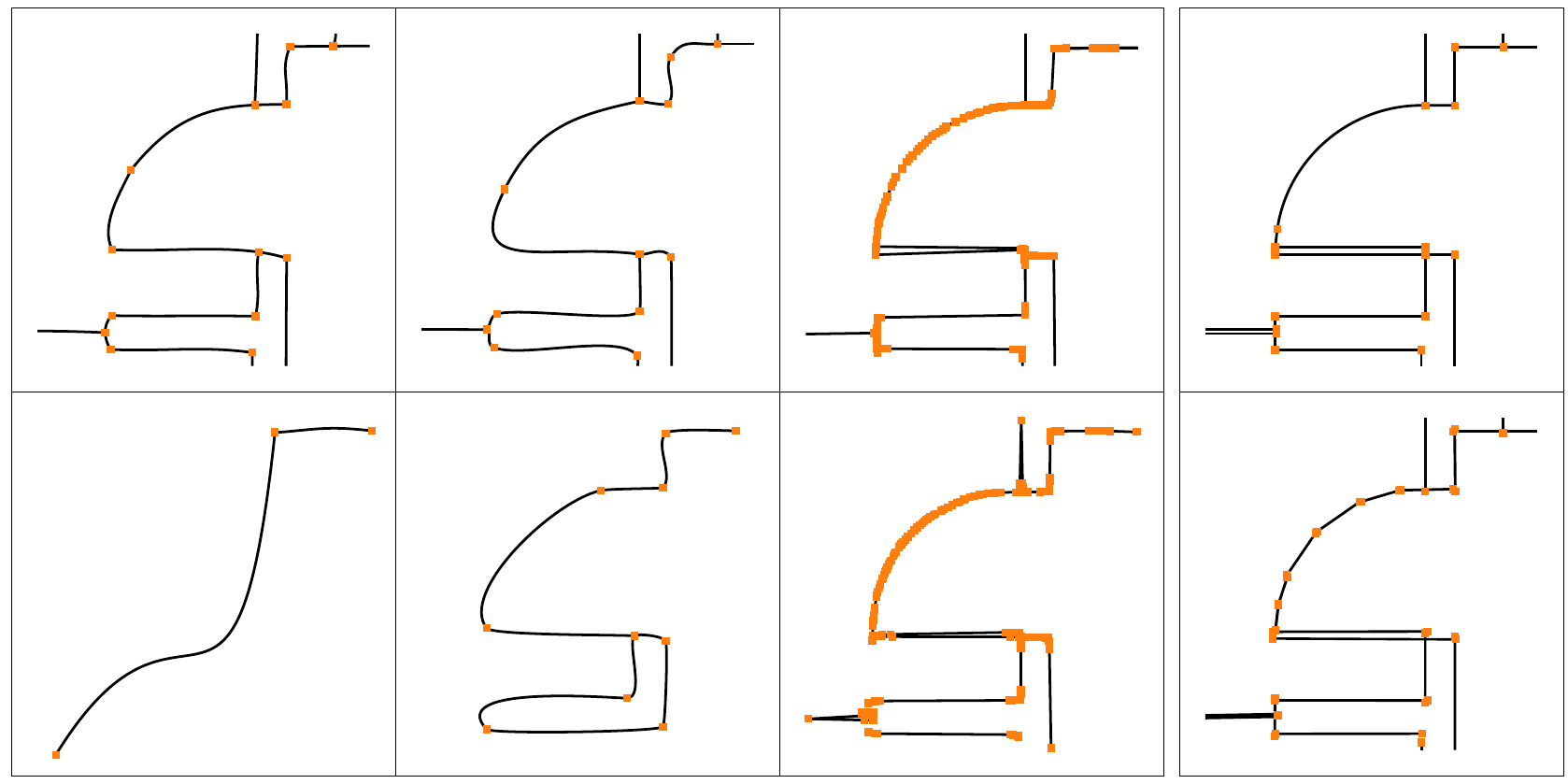}};
                    \begin{scope}[x={(image.south east)},y={(image.north west)}]
                        \footnotesize
                        \def\y{-1.5ex}
                        \def\dx{.125}
                        \node at (\dx, \y)   [align=center] {FvS~\cite{favreau2016fidelity}};
                        \node at (3*\dx, \y) [align=center] {CHD~\cite{donati2019complete}};
                        \node at (5*\dx, \y) [align=center] {PVF~\cite{bessmeltsev2019vectorization}};
                        \node at (1 - \dx, \y) [align=center] {GT / Our method};
                        \node[rotate=90] at (\y, .25) {On patch};
                        \node[rotate=90] at (\y, .75) {On full image};
                    \end{scope}
    \end{tikzpicture}}
    \vspace{.25cm}
    \centerline{\begin{tikzpicture}
                    \node[anchor=south west,inner sep=0] (image) at (0,0) {\includegraphics[width=\linewidth]{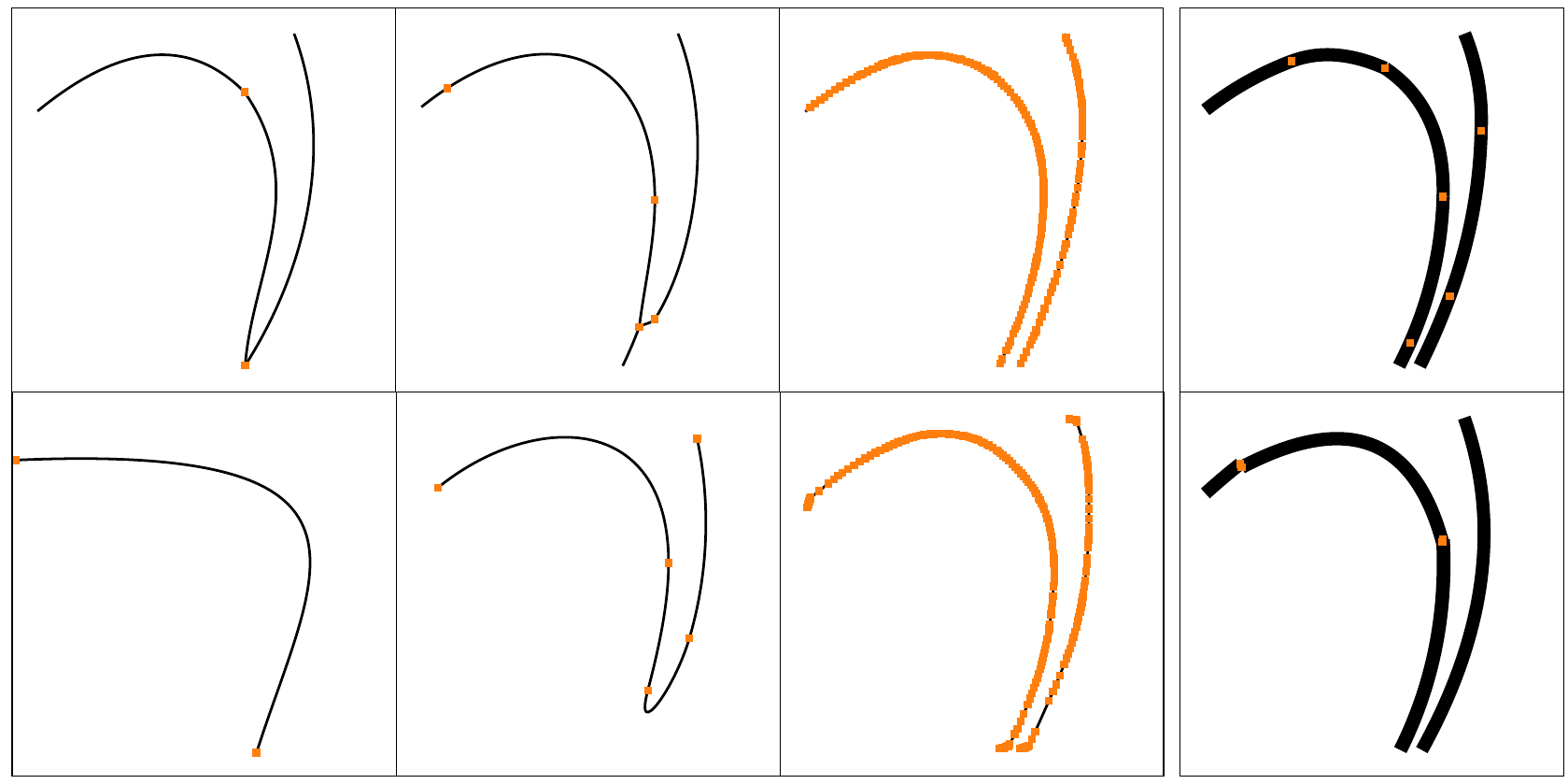}};
                    \begin{scope}[x={(image.south east)},y={(image.north west)}]
                        \footnotesize
                        \def\y{-1.5ex}
                        \def\dx{.125}
                        \node at (\dx, \y)   [align=center] {FvS~\cite{favreau2016fidelity}};
                        \node at (3*\dx, \y) [align=center] {CHD~\cite{donati2019complete}};
                        \node at (5*\dx, \y) [align=center] {PVF~\cite{bessmeltsev2019vectorization}};
                        \node at (1 - \dx, \y) [align=center] {GT / Our method};
                        \node[rotate=90] at (\y, .25) {On patch};
                        \node[rotate=90] at (\y, .75) {On full image};
                    \end{scope}
    \end{tikzpicture}}
    \caption{Results of the prior work on small patches and
    the respective patches cut from the results on whole images.
    Endpoints of the primitives are shown in orange.
    The whole images are shown at the top of Figure~\ref{fig:pfp-sup} and in Figure 6 from the main text.
    }
    \label{fig:prior-on-patches}
\end{figure*}

The main steps of our vectorization system, the primitive extraction network and refinement,
operate on small patches of the image, while the methods that we compare with operate on whole images.
To demonstrate that our method outperforms these ones not only because of this divide-and-merge strategy,
in Figure~\ref{fig:prior-on-patches} we show example outputs of these methods applied to small patches
in comparison to the respective patches cut from the results on whole images.

The methods of~\cite{bessmeltsev2019vectorization,donati2019complete} produce similar results
on small patches and whole images, as expected since they use local operations.
The method of~\cite{favreau2016fidelity} produces worse results on patches.

\section{Generalization to cartoon drawings}
\label{sec:sup.cartoon}
\begingroup
Figure~\ref{fig:cartoon} shows the results produced by our system on clean cartoon drawings.
Here we used the version operating on curves, with the neural networks trained on technical drawings.

Our system produces reasonable results,
although the predictions of the primitive extraction network are qualitatively less accurate
than in case of technical drawings that we focused on.
A proper extension of our system to a different kind of drawings would require
(1) the corresponding training dataset for the primitive extraction network,
and (2) in case of rough sketches,
either a proper training dataset with clean targets for the preprocessing cleaning step,
or significant changes of the refinement step.

\newcommand\myfigure[2][]{%
\begingroup
\setkeys{myfigure}{#1}%
\centerline{%
\begin{tikzpicture}
    \node[anchor=south west,inner sep=0] (image) at (0,0) {\includegraphics[width=\linewidth]{\image}};
    \begin{scope}[x={(image.south east)},y={(image.north west)}]
        \footnotesize
        \def\y{-0.75\baselineskip}
        \def\dx{1./8}
        % Guide coordinate grid
        \ifgrid
        \begingroup
        \fontsize{2}{2}\selectfont
        \draw[help lines,red,xstep=.01,ystep=.01] (0,0) grid (1,1);
        \foreach \x in {0,1,...,99} { \node [anchor=north] at (\x/100,0) {\x}; }
        \foreach \y in {0,1,...,99} { \node [anchor=east] at (0,\y/100) {\y}; }
        \endgroup
        \fi
        % Captions
        \node at ($(\dx,\y) + (0,.25)$) [align=center] {Input};
        \node at (3*\dx, \y) [align=center] {NN};
        \node at (5*\dx, \y) [align=center] {NN + Refinement};
        \node at (7*\dx, \y) [align=center] {Full};
        % Size
        \begingroup
        \def\armg{.002}
        \fontsize{6}{6}\selectfont
        \draw[latex-latex] (1.005, \splity + \armg) -- (1.005, 1 - \armg);
        \node[rotate=-90] at ($(1.005,0) + (0,\splity)!.5!(0,1) + (.75\baselineskip,0)$) {\fullsize};
        \endgroup
    \end{scope}
\end{tikzpicture}%
}%
\endgroup
}
\makeatletter
\define@boolkey{myfigure}[]{grid}[false]{}
\define@cmdkey{myfigure}[]{image}{}
\define@cmdkey{myfigure}[]{splity}{}
\define@cmdkey{myfigure}[]{fullsize}{}

\begin{figure*}[h!]
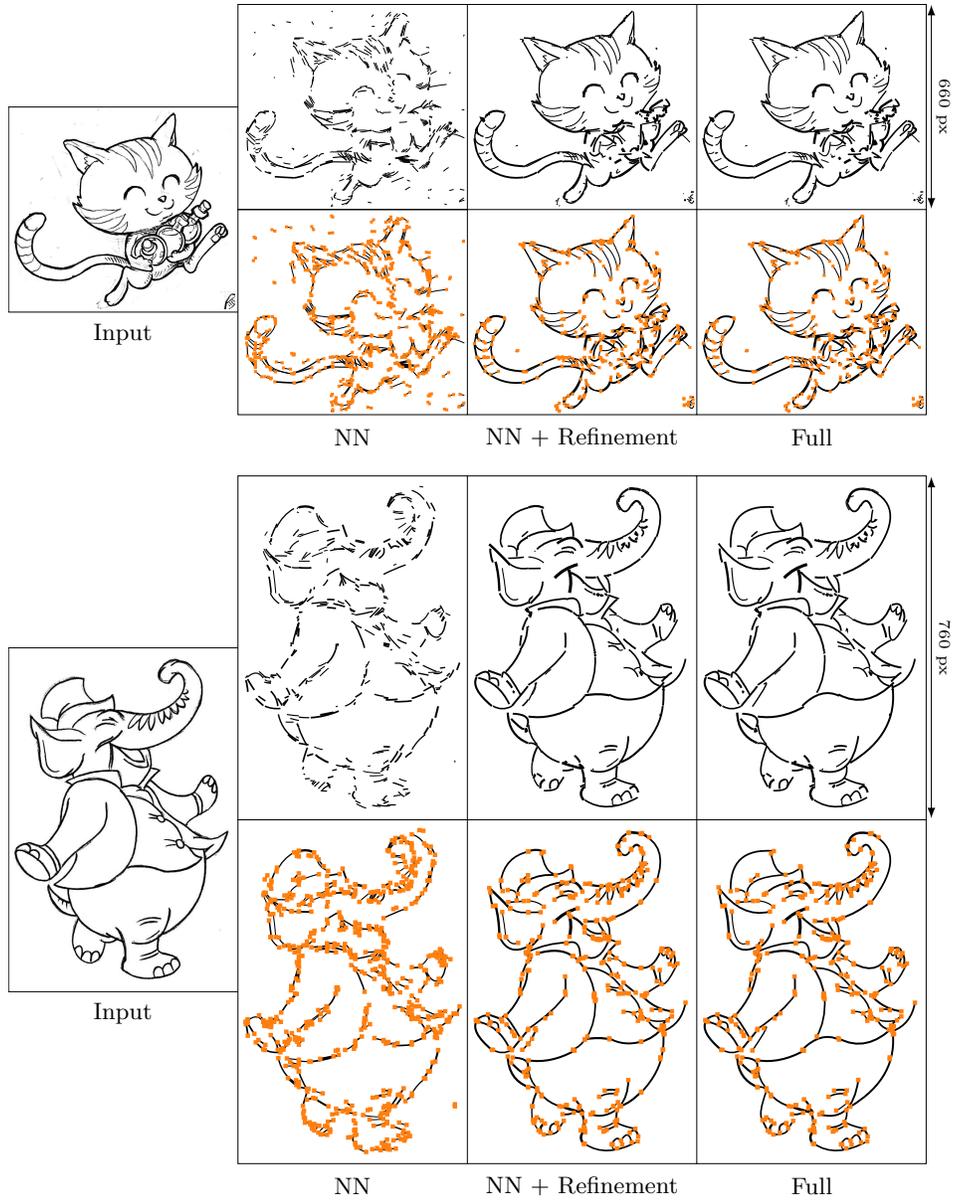

    \myfigure[image={supp/cartoon/kitty}, splity=.5, fullsize=660 px]{}
    \vspace{.25cm}
    \myfigure[image={supp/cartoon/elephant}, splity=.5, fullsize=760 px]{}
    \caption{Qualitative results of our system on clean cartoon drawings.
    Endpoints of primitives are shown in orange. The input image on the top is copyrighted by David Revoy www.davidrevoy.com  under CC-by 4.0 license and on the bottom from www.easy-drawings-and-sketches.com, \textsuperscript{\textcopyright} Ivan Huska.}
    \label{fig:cartoon}
\end{figure*}
\endgroup

\section{Additional results}
\label{sec:aresults}
\begingroup
In this section, we show more qualitative comparisons on test set for both PFP in Figure~\ref{fig:pfp-sup} and ABC in Figure~\ref{fig:abc-sup} datasets and on real data in Figure~\ref{fig:real-sup}.

\newcommand\myfigure[2][]{%
\begingroup
\setkeys{myfigure}{#1}%
\centerline{%
\begin{tikzpicture}
    \node[anchor=south west,inner sep=0] (image) at (0,0) {\includegraphics[width=\linewidth]{\image}};
    \begin{scope}[x={(image.south east)},y={(image.north west)}]
        \footnotesize
        \def\y{-1.75\baselineskip}
        \def\dx{.1}
        \ifgrid%
        % Guide coordinate grid
        \begingroup
        \fontsize{2}{2}\selectfont
        \draw[help lines,red,xstep=.01,ystep=.01] (0,0) grid (1,1);
        \foreach \x in {0,1,...,99} { \node [anchor=north] at (\x/100,0) {\x}; }
        \foreach \y in {0,1,...,99} { \node [anchor=east] at (0,\y/100) {\y}; }
        \endgroup
        \fi
        % Captions
        \node at (\dx, \y)   [align=center] {FvS~\cite{favreau2016fidelity}          \\ \fvsiou / \fvsdh  \\ \fvsdm / \fvsP};
        \node at (3*\dx, \y) [align=center] {CHD~\cite{donati2019complete}           \\ \chdiou / \chddh  \\ \chddm / \chdP};
        \node at (5*\dx, \y) [align=center] {PVF~\cite{bessmeltsev2019vectorization} \\ \pvfiou / \pvfdh  \\ \pvfdm / \pvfP};
        \node at (7*\dx, \y) [align=center] {Our method                              \\ \ouriou / \ourdh  \\ \ourdm / \ourP};
        \node at (9*\dx, \y) [align=center] {Ground truth, \\ \numberofprimitives\ \gtP};
        % Size
        \begingroup
        \def\armg{.002}
        \fontsize{6}{6}\selectfont
        \draw[latex-latex] (1.005, \splity + \armg) -- (1.005, 1 - \armg);
        \node[rotate=-90] at ($(1.005,0) + (0,\splity)!.5!(0,1) + (.75\baselineskip,0)$) {\fullsize};
        \draw[latex-latex] (1.005, 0 + \armg) -- (1.005, \splity - \armg);
        \node[rotate=-90] at ($(1.005,0) + (0,0)!.5!(0,\splity) + (.75\baselineskip,0)$) {\closesize};
        \endgroup
    \end{scope}
\end{tikzpicture}%
}%
\endgroup
}
\makeatletter
\define@boolkey{myfigure}[]{grid}[false]{}
\define@cmdkey{myfigure}[]{image}{}
\define@cmdkey{myfigure}[]{fvsiou}{}
\define@cmdkey{myfigure}[]{fvsdh}{}
\define@cmdkey{myfigure}[]{fvsdm}{}
\define@cmdkey{myfigure}[]{fvsP}{}
\define@cmdkey{myfigure}[]{chdiou}{}
\define@cmdkey{myfigure}[]{chddh}{}
\define@cmdkey{myfigure}[]{chddm}{}
\define@cmdkey{myfigure}[]{chdP}{}
\define@cmdkey{myfigure}[]{pvfiou}{}
\define@cmdkey{myfigure}[]{pvfdh}{}
\define@cmdkey{myfigure}[]{pvfdm}{}
\define@cmdkey{myfigure}[]{pvfP}{}
\define@cmdkey{myfigure}[]{ouriou}{}
\define@cmdkey{myfigure}[]{ourdh}{}
\define@cmdkey{myfigure}[]{ourdm}{}
\define@cmdkey{myfigure}[]{ourP}{}
\define@cmdkey{myfigure}[]{gtP}{}
\define@cmdkey{myfigure}[]{splity}{}
\define@cmdkey{myfigure}[]{fullsize}{}
\define@cmdkey{myfigure}[]{closesize}{}
\makeatother

\begin{figure*}[h!]
    \myfigure[%
    image={supp/pfp_1},%
    fvsiou={31\%}, fvsdh={99px}, fvsdm={0.9px}, fvsP={\textbf{1098}},%
    chdiou={25\%}, chddh={178px}, chddm={1.4px}, chdP={1679},%
    pvfiou={64\%}, pvfdh={\textbf{46px}}, pvfdm={0.7px}, pvfP={50k},%
    ouriou={\textbf{89\%}}, ourdh={\textbf{46px}}, ourdm={\textbf{0.3px}}, ourP={1513},%
    gtP=2238,%
    splity=.58, fullsize=1987 px, closesize=432 px]{}
    \vspace{.25cm}
    \myfigure[%
    image={supp/pfp_2},%
    fvsiou={33\%}, fvsdh={398px}, fvsdm={1.6px}, fvsP={\textbf{907}},%
    chdiou={21\%}, chddh={141px}, chddm={1.3px}, chdP={1808},%
    pvfiou={71\%}, pvfdh={97px}, pvfdm={\textbf{0.4px}}, pvfP={53k},%
    ouriou={\textbf{80\%}}, ourdh={\textbf{25px}}, ourdm={\textbf{0.4px}}, ourP={2336},%
    gtP=3069,%
    splity=.505, fullsize=2025 px, closesize=343 px]{}
    \caption{Qualitative comparison on PFP images,
    and values of metrics IoU / $\mathrm{\hausdorff}$ / $\mathrm{\meanmin}$ / \numberofprimitives\ with best in bold.
    Endpoints of the primitives are shown in orange.}
    \label{fig:pfp-sup}
\end{figure*}

\begin{figure*}[h!]
    \myfigure[
    image={supp/abc_2},
    fvsiou={86\%},          fvsdh={23px},         fvsdm={1.3px},          fvsP={\textbf{88}},
    chdiou={68\%},          chddh={5px},          chddm={1.0px},          chdP={211},
    pvfiou={\textbf{96\%}}, pvfdh={\textbf{2px}}, pvfdm={\textbf{0.2px}}, pvfP={\color{red}{20k}},
    ouriou={82\%},          ourdh={14px},         ourdm={\textbf{0.2px}}, ourP={147},
    gtP=122,
    splity=.495, fullsize=2604 px, closesize=536 px]{}
    \vspace{.25cm}
    \myfigure[
    image={supp/abc_3},
    fvsiou={74\%},          fvsdh={27px},          fvsdm={0.7px},          fvsP={\textbf{433}},
    chdiou={64\%},          chddh={17px},          chddm={0.9px},          chdP={994},
    pvfiou={\textbf{91\%}}, pvfdh={\textbf{6px}},  pvfdm={\textbf{0.3px}}, pvfP={\color{red}{43k}},
    ouriou={76\%},          ourdh={34px},          ourdm={0.6px},          ourP={579},
    gtP=973,
    splity=.52, fullsize=3303 px, closesize=549 px]{}
    \caption{Qualitative comparison on ABC images,
    and values of  IoU / $\mathrm{\hausdorff}$ / $\mathrm{\meanmin}$ / \numberofprimitives\ metrics, with the best result in boldface.
    The endpoints of  primitives are shown in orange.}
    \label{fig:abc-sup}
\end{figure*}

\begin{figure*}[t]
    \centerline{\begin{tikzpicture}
                    \tiny
                    \node[anchor=south west,inner sep=0] (image) at (0,0) {\includegraphics[width=\linewidth]{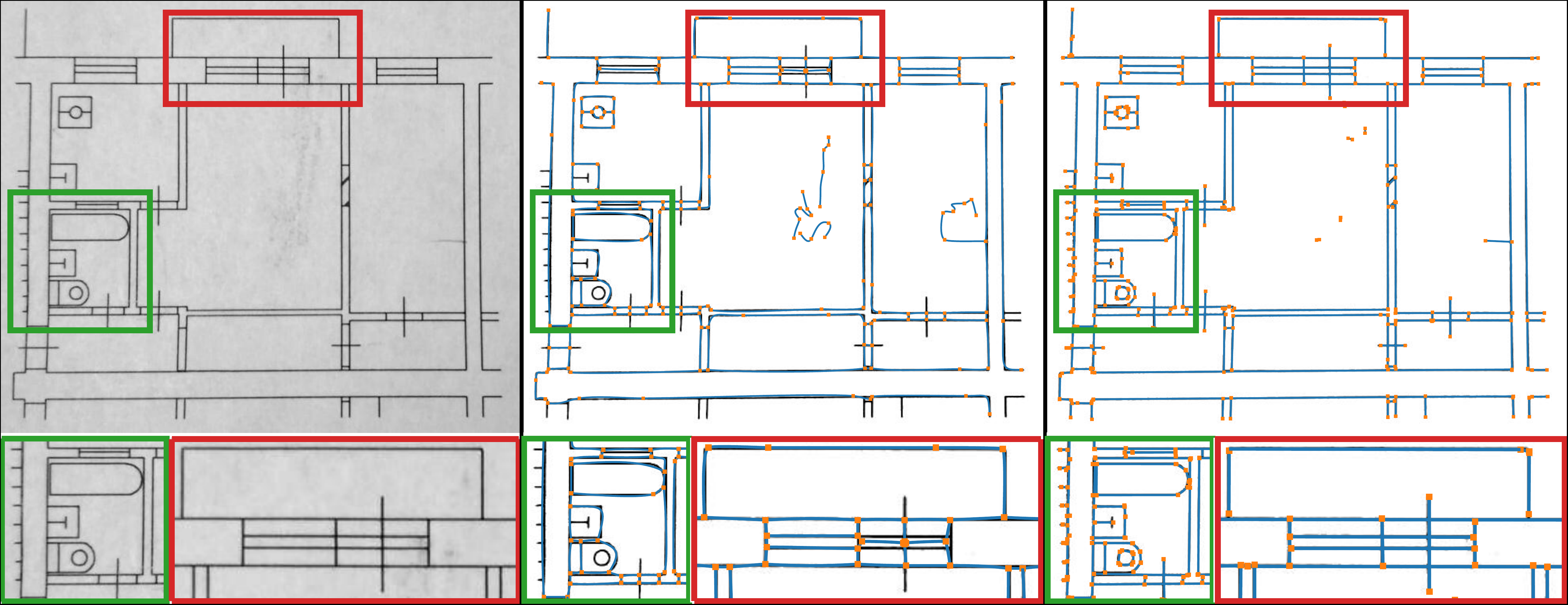}};
                    \begin{scope}[x={(image.south east)},y={(image.north west)}]
                        \footnotesize
                        \def\y{-0.75\baselineskip}
                        \def\dx{1./6}
                        % Captions
                        \node at (\dx, \y) {Input image};
                        \node at (3*\dx, \y)     {CHD~\cite{donati2019complete}, 38\%          / 230};
                        \node at (5*\dx, \y) {Our method,                    \textbf{81\%} / \textbf{187} };
                        % Size
                        \begingroup
                        \def\armg{.002}
                        \fontsize{6}{6}\selectfont
                        \def\splity{.28}
                        \def\fullsize{548 px}
                        \def\closesize{116 px}
                        \draw[latex-latex] (1.005, \splity + \armg) -- (1.005, 1 - \armg);
                        \node[rotate=-90] at ($(1.005,0) + (0,\splity)!.5!(0,1) + (.75\baselineskip,0)$) {\fullsize};
                        \draw[latex-latex] (1.005, 0 + \armg) -- (1.005, \splity - \armg);
                        \node[rotate=-90] at ($(1.005,0) + (0,0)!.5!(0,\splity) + (.75\baselineskip,0)$) {\closesize};
                        \endgroup
                    \end{scope}
    \end{tikzpicture}}
    \vspace{.25cm}
    \centerline{\begin{tikzpicture}
                    \tiny
                    \node[anchor=south west,inner sep=0] (image) at (0,0) {\includegraphics[width=\linewidth]{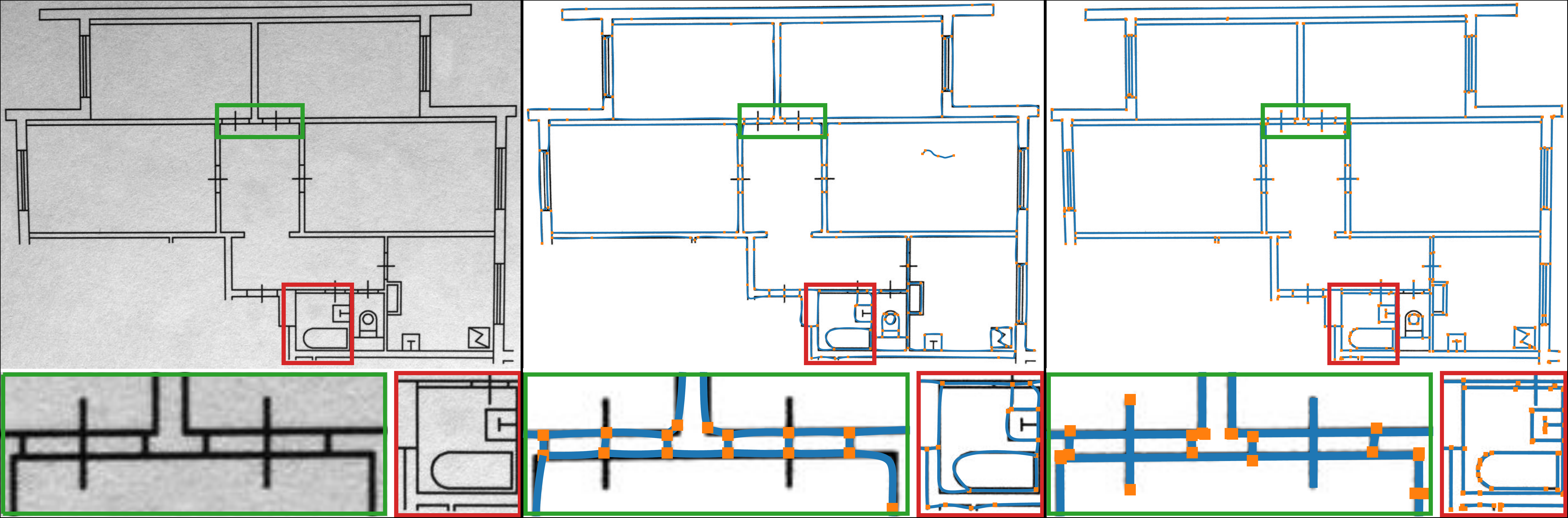}};
                    \begin{scope}[x={(image.south east)},y={(image.north west)}]
                        \footnotesize
                        \def\y{-0.75\baselineskip}
                        \def\dx{1./6}
                        % Captions
                        \node at (\dx, \y) {Input image};
                        \node at (3*\dx, \y)     {CHD~\cite{donati2019complete}, 44\%          / 226};
                        \node at (5*\dx, \y) {Our method,                    \textbf{84\%} / \textbf{174} };
                        % Size
                        \begingroup
                        \def\armg{.002}
                        \fontsize{6}{6}\selectfont
                        \def\splity{.28}
                        \def\fullsize{658 px}
                        \def\closesize{140 px}
                        \draw[latex-latex] (1.005, \splity + \armg) -- (1.005, 1 - \armg);
                        \node[rotate=-90] at ($(1.005,0) + (0,\splity)!.5!(0,1) + (.75\baselineskip,0)$) {\fullsize};
                        \draw[latex-latex] (1.005, 0 + \armg) -- (1.005, \splity - \armg);
                        \node[rotate=-90] at ($(1.005,0) + (0,0)!.5!(0,\splity) + (.75\baselineskip,0)$) {\closesize};
                        \endgroup
                    \end{scope}
    \end{tikzpicture}}
    \caption{Qualitative comparison on real noisy images, and values of metric IoU / \numberofprimitives\ with best in bold.
    Primitives are shown in blue with the endpoints in orange on top of the cleaned raster image.}
    \label{fig:real-sup}
\end{figure*}
\endgroup

\section{Qualitative ablation study}
\label{sec:ablstudy}
\begingroup
In this section, we show qualitative results obtained using our system with the (a) full model without refinement and post-processing steps, (b) full model without post-processing, (c) full model. You can see this comparison on ABC dataset in Figure~\ref{fig:abl-sup1} and Figure~\ref{fig:abl-sup2}.

\newcommand\myfigure[2][]{%
\begingroup
\setkeys{myfigure}{#1}%
\centerline{%
\begin{tikzpicture}
    \node[anchor=south west,inner sep=0] (image) at (0,0) {\includegraphics[width=\linewidth]{\image}};
    \begin{scope}[x={(image.south east)},y={(image.north west)}]
        \footnotesize
        \def\y{-1.75\baselineskip}
        \def\dx{1./8}
        \ifgrid%
        % Guide coordinate grid
        \begingroup
        \fontsize{2}{2}\selectfont
        \draw[help lines,red,xstep=.01,ystep=.01] (0,0) grid (1,1);
        \foreach \x in {0,1,...,99} { \node [anchor=north] at (\x/100,0) {\x}; }
        \foreach \y in {0,1,...,99} { \node [anchor=east] at (0,\y/100) {\y}; }
        \endgroup
        \fi
        % Captions
        \node at (\dx, \y)   [align=center] {NN               \\ \nniou  / \nndh  \\ \nndm  / \nnP };
        \node at (3*\dx, \y) [align=center] {NN + Refinement  \\ \refiou / \refdh \\ \refdm / \refP };
        \node at (5*\dx, \y) [align=center] {Full             \\ \fuliou / \fuldh \\ \fuldm / \fulP };
        \node at (7*\dx, \y) [align=center] {Ground truth, \\ \numberofprimitives\ \gtP};
        % Size
        \begingroup
        \def\armg{.002}
        \fontsize{6}{6}\selectfont
        \draw[latex-latex] (1.005, \splity + \armg) -- (1.005, 1 - \armg);
        \node[rotate=-90] at ($(1.005,0) + (0,\splity)!.5!(0,1) + (.75\baselineskip,0)$) {\fullsize};
        \endgroup
    \end{scope}
\end{tikzpicture}%
}%
\endgroup
}
\makeatletter
\define@boolkey{myfigure}[]{grid}[false]{}
\define@cmdkey{myfigure}[]{image}{}
\define@cmdkey{myfigure}[]{nniou}{}
\define@cmdkey{myfigure}[]{nndh}{}
\define@cmdkey{myfigure}[]{nndm}{}
\define@cmdkey{myfigure}[]{nnP}{}
\define@cmdkey{myfigure}[]{refiou}{}
\define@cmdkey{myfigure}[]{refdh}{}
\define@cmdkey{myfigure}[]{refdm}{}
\define@cmdkey{myfigure}[]{refP}{}
\define@cmdkey{myfigure}[]{fuliou}{}
\define@cmdkey{myfigure}[]{fuldh}{}
\define@cmdkey{myfigure}[]{fuldm}{}
\define@cmdkey{myfigure}[]{fulP}{}
\define@cmdkey{myfigure}[]{gtP}{}
\define@cmdkey{myfigure}[]{splity}{}
\define@cmdkey{myfigure}[]{fullsize}{}
\makeatother

\begin{figure*}[h!]
    \myfigure[
    image={supp/abl_2},
    nniou={51\%},           nndh={24px},           nndm={1.0px},           nnP={265},
    refiou={\textbf{91\%}}, refdh={\textbf{19px}}, refdm={\textbf{0.2px}}, refP={232},
    fuliou={85\%},          fuldh={\textbf{19px}}, fuldm={0.3px},          fulP={\textbf{84}},
    gtP=89, splity=.5, fullsize=1220 px]{}
    \caption{Qualitative comparison on ABC images,
    and values of  IoU / $\mathrm{\hausdorff}$ / $\mathrm{\meanmin}$ / \numberofprimitives\  metrics, with the best results shown in boldface.
    The endpoints of primitives are shown in orange.}
    \label{fig:abl-sup1}
\end{figure*}

\begin{figure*}[h!]
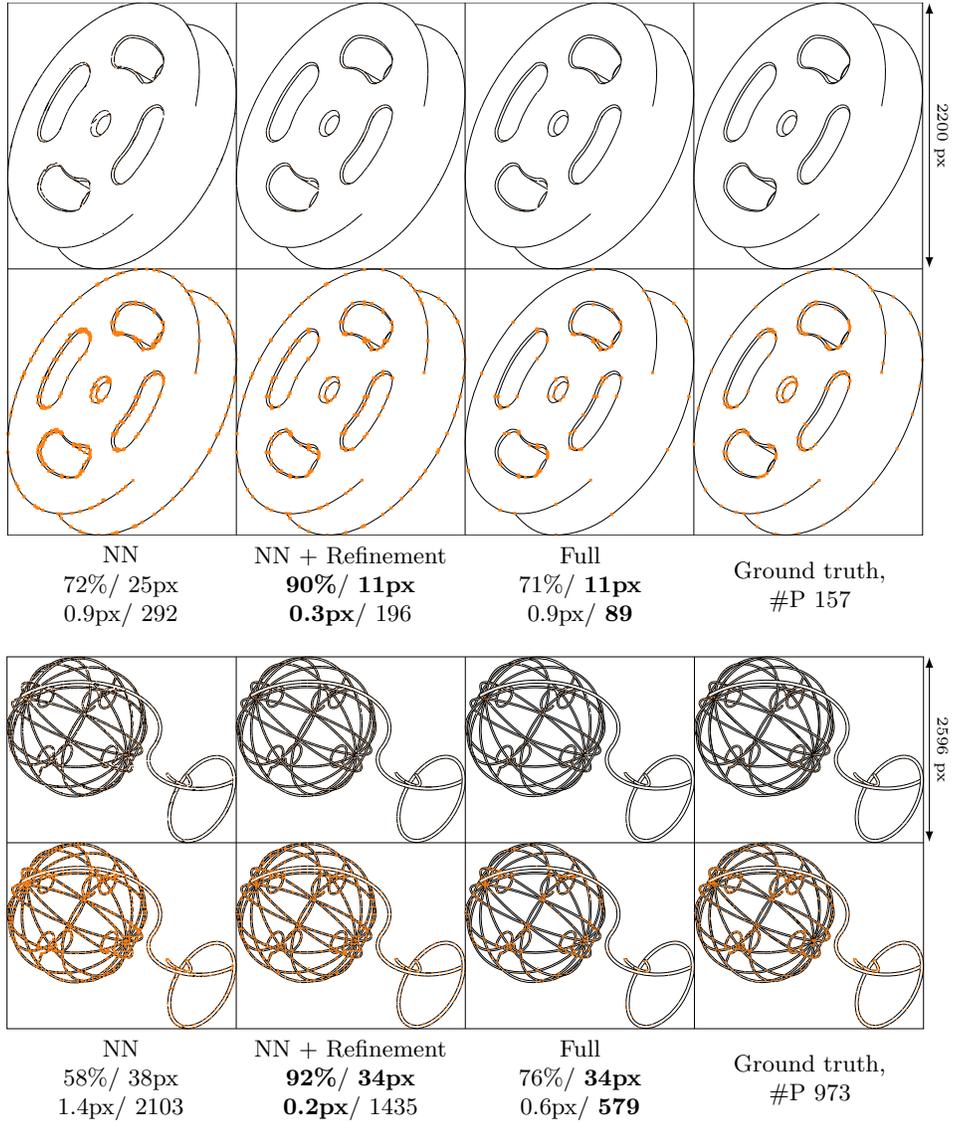

    \myfigure[
    image={supp/abl_3},
    nniou={72\%},           nndh={25px},           nndm={0.9px},           nnP={292},
    refiou={\textbf{90\%}}, refdh={\textbf{11px}}, refdm={\textbf{0.3px}}, refP={196},
    fuliou={71\%},          fuldh={\textbf{11px}}, fuldm={0.9px},          fulP={\textbf{89}},
    gtP=157, splity=.5, fullsize=2200 px]{}
    \vspace{.25cm}
    \myfigure[
    image={supp/abl_4},
    nniou={58\%},           nndh={38px},           nndm={1.4px},           nnP={2103},
    refiou={\textbf{92\%}}, refdh={\textbf{34px}}, refdm={\textbf{0.2px}}, refP={1435},
    fuliou={76\%},          fuldh={\textbf{34px}}, fuldm={0.6px},          fulP={\textbf{579}},
    gtP=973, splity=.5, fullsize=2596 px]{}
    \caption{Qualitative comparison on ABC images,
    and values of metrics IoU / $\mathrm{\hausdorff}$ / $\mathrm{\meanmin}$ / \numberofprimitives\ with best in bold.
    Endpoints of the primitives are shown in orange.}
    \label{fig:abl-sup2}
\end{figure*}
\endgroup

\FloatBarrier
\section{Details on refinement algorithm}
\label{sec:refalg}
\subsection{Overall idea}

The underlying idea in our approach is to use interaction potentials, qualitatively similar, \eg., to electrostatic interaction,  to construct our optimization functionals.  Fixed charges are associated with filled pixels, and moving charges to the points on primitives.    Primitives and filled pixels of the raster image are assigned charges of different signs: negative for pixels and positive for primitives.  As a consequence, primitives and filled pixels  and are attracted, and primitives repulse other primitives.  Internal charges push primitives to expand, because their internal charges are repulsing each other.  A number of modifications need to be made to this general approach to avoid undesirable minima.

The interaction energy of two charges at points \(\coordinatevector_1, \coordinatevector_2\) is given by
\begin{equation}
    \label{eq:pointpot}
    \charge_1 \charge_2 \drpotential{\norm{\coordinatevector_1 - \coordinatevector_2}},
\end{equation}
where \(\charge_1,\charge_2\) are signed charges, and  \(\drpotential{\coordinatevectorlen}\) is the interaction potential of two charges at the distance \(\coordinatevectorlen\) from each other. 
The standard 3D electrostatic potential is \(\frac{1}{\coordinatevectorlen}\); we replace it by
 an exponentially decaying potential as explained at the end of this section. 
The total energy is obtained by summation/integration over all charge pairs.

\subsubsection{Energy.}
We split our energy into three parts: primitive-pixel interactions,\linebreak[100] interactions between distinct primitives and interaction between charges inside the same primitive. As the charges at pixels do not move, their interactions with each other can be ignored. 

\begin{equation}
    \label{eq:efull}
    \edr = \sum\limits_{\kprim,\ipix} \edrprimpix_{\kprim,\ipix} + \sum_{\kprim < \jprim} \edrprimprim_{\kprim,\jprim} + \sum_{\kprim} \edrprim_{\kprim}.
\end{equation}

Three parts of the energy have the following form:
\begin{equation}
    \label{eq:eprimpix}
    \edrprimpix_{\kprim,\ipix} = -\chargeraster \iint\limits_{\primitivearea_{\kprim}} \drpotential{\norm{\coordinatevector - \coordinatevector_{\ipix}}} \dprimitivearea ,
\end{equation}
where \(\chargeraster\) is the pixel intensity,  \(\coordinatevector_{\ipix}\) and
\(\primitivearea_{\kprim}\) domain covered by the primitive; 

\begin{equation}
    \label{eq:eprimprim}
    \edrprimprim_{\kprim,\jprim} = \iint\limits_{\primitivearea_{\kprim}} \iint\limits_{\primitivearea_{\jprim}} \drpotential{\norm{\coordinatevector_1 - \coordinatevector_2}} \dprimitivearea{1}\dprimitivearea{2};
\end{equation}
and
\begin{equation}
    \label{eq:eprim}
    \edrprim_{\kprim} = \frac{1}{2}\edrprimprim_{\kprim,\kprim} = \frac{1}{2} \iint\limits_{\primitivearea_{\kprim}} \iint\limits_{\primitivearea_{\kprim}} \drpotential{\norm{\coordinatevector_1 - \coordinatevector_2}} \dprimitivearea{1}\dprimitivearea{2}.
\end{equation}
\paragraph*{Energy properties.} Observe that pixel-primitive interaction is negative and decays (increases in magnitude) as primitive get close to a pixel, and also decreases as a primitive increase in size (more coverage is good).  Primitive-primitive interaction energy is positive, decreases as the primitives move apart and also as the size of the primitives decreases.  Finally, the self-interaction energy of a primitive is positive, does not depend on the primitive position and decreases if the primitive shrinks. 

\subsection{Mean-field-based optimization}

For optimizing the energy efficiently, we use an approach based on a standard approach in the \emph{mean-field} theory: the interactions between particles are viewed as individual interactions with a mean field, which is then updated using updated particle positions.

The basic gradient descent update of
\(\parameterindex^{\text{th}}\) parameter of \(\primitiveindex^{\text{th}}\) primitive is:
\begin{equation}
    \primparameter_{\primitiveindex,\parameterindex} \leftarrow
    \primparameter_{\primitiveindex,\parameterindex} - \lambda
    \frac{\partial \edr}{\partial \primparameter_{\primitiveindex,\parameterindex}}.
\end{equation}

We split the primitive parameters into size and position parameters and spell out the derivatives explicitly in each case, highlighting in blue the parts of the expressions that depend on the primitive. 

\begin{equation}
\begin{gathered}
    \frac{\partial}{\partial \primparameter_{\primitiveindex,\parameterindex}^{\braces{\poslabel,\sizelabel}}}
    \sum_{\kprim,\ipix} \edrprimpix_{\kprim,\ipix} =
    \partial \sum_{\ipix} \edrprimpix_{\primitiveindex,\ipix} = \\
    -\sum_{\ipix}\chargeraster \partial \iint\limits_{\optimized{\primitivearea_{\primitiveindex}}}
    \drpotential{\norm{\coordinatevector - \coordinatevector_{\ipix}}} \dprimitivearea ,
\end{gathered}
\end{equation}

\begin{equation}
\begin{gathered}
    \frac{\partial}{\partial \primparameter_{\primitiveindex,\parameterindex}^{\braces{\poslabel,\sizelabel}}}
    \sum_{\kprim < \jprim} \edrprimprim_{\kprim,\jprim} =
    \partial \sum_{\jprim\ne\primitiveindex} \edrprimprim_{\primitiveindex,\jprim} = \\
    \partial \iint\limits_{\optimized{\primitivearea_{\primitiveindex}}}\sum_{\jprim\ne\primitiveindex} \iint\limits_{\primitivearea_{\jprim}}
    \drpotential{\norm{\coordinatevector_1 - \coordinatevector_2}} \dprimitivearea{1}\dprimitivearea{2},
\end{gathered}
\end{equation}

\begin{equation}
    \frac{\partial}{\partial \primparameter_{\primitiveindex,\parameterindex}^{\poslabel}}
    \sum_{\kprim} \edrprim_{\kprim} = 0,
\end{equation}

\begin{equation}
\begin{gathered}
    \frac{\partial}{\partial \primparameter_{\primitiveindex,\parameterindex}^{\sizelabel}}
    \sum_{\kprim} \edrprim_{\kprim} =
    \partial \edrprim_{\primitiveindex} = \\
    \frac{1}{2}\partial \iint\limits_{\optimized{\primitivearea_{\primitiveindex}}} \iint\limits_{\primitivearea_{\primitiveindex}}
    \drpotential{\norm{\coordinatevector_1 - \coordinatevector_2}} \dprimitivearea{1}\dprimitivearea{2} +
    \frac{1}{2}\partial \iint\limits_{\primitivearea_{\primitiveindex}} \iint\limits_{\optimized{\primitivearea_{\primitiveindex}}}
    \drpotential{\norm{\coordinatevector_1 - \coordinatevector_2}} \dprimitivearea{1}\dprimitivearea{2} = \\
    \partial \iint\limits_{\optimized{\primitivearea_{\primitiveindex}}} \iint\limits_{\primitivearea_{\primitiveindex}}
    \drpotential{\norm{\coordinatevector_1 - \coordinatevector_2}} \dprimitivearea{1}\dprimitivearea{2},
\end{gathered}
\end{equation}

The complete expressions for the energy derivatives with respect to positional parameters are: 
\begin{equation}
\begin{gathered}
    \frac{\partial \edr}{\partial \primparameter_{\primitiveindex,\parameterindex}^{\poslabel}} =
    \partial \sum_{\ipix} \edrprimpix_{\primitiveindex,\ipix} +
    \partial \sum_{\jprim\ne\primitiveindex} \edrprimprim_{\primitiveindex,\jprim} = \\
    \partial \iint\limits_{\optimized{\primitivearea_{\primitiveindex}}} \brackets{
        \sum_{\jprim\ne\primitiveindex}\iint\limits_{\primitivearea_{\jprim}} \drpotential{\norm{\coordinatevector - \coordinatevector_1}} \dprimitivearea{1} - 
        \sum_{\ipix}\chargeraster \drpotential{\norm{\coordinatevector - \coordinatevector_{\ipix}}}
    } \dprimitivearea.
\end{gathered}
\end{equation}
For size parameters, we obtain the following expression 
\begin{equation}
\begin{gathered}
    \frac{\partial \edr}{\partial \primparameter_{\primitiveindex,\parameterindex}^{\sizelabel}} =
    \partial \sum_{\ipix} \edrprimpix_{\primitiveindex,\ipix} +
    \partial \sum_{\jprim} \edrprimprim_{\primitiveindex,\jprim} = \\
    \partial \iint\limits_{\optimized{\primitivearea_{\primitiveindex}}} \brackets{
        \sum_{\jprim}\iint\limits_{\primitivearea_{\jprim}} \drpotential{\norm{\coordinatevector - \coordinatevector_1}} \dprimitivearea{1} - 
        \sum_{\ipix}\chargeraster \drpotential{\norm{\coordinatevector - \coordinatevector_{\ipix}}}
    } \dprimitivearea,
\end{gathered}
\end{equation}
where  \(\jprim\) ranges over all primitives including \(\primitiveindex\)

We can interpret these derivatives as derivatives of a different function
\begin{equation}
    \label{eq:fullpotential}
    \edr^* = \sum_{\primitiveindex} \edrpos + \edrsize.
\end{equation}
with terms defined below. Each term corresponds to particular parameters of one of the primitives, 
and can be viewed as the interaction energy of the primitive with a background charge distribution defined by all primitives at a given instance in time.   

\begin{equation}
    \label{eq:Ei}
    \edrpointprim\parens{\charge} = \iint\limits_{\canvasarea} \charge\parens{\coordinatevector_1}
    \iint\limits_{\primitivearea_{\primitiveindex}}
    \drpotential{\norm{\coordinatevector_1 - \coordinatevector_2}} \dprimitivearea{2}\dprimitivearea{1},
\end{equation}
\begin{equation}
    \edrpos = \withcondition{\edrpointprim\parens{\charge^\poslabel_\primitiveindex}}{\primitiveparametersetsize_\primitiveindex = \mathrm{const}},\quad
    \edrsize = \withcondition{\edrpointprim\parens{\charge^\sizelabel_\primitiveindex}}{\primitiveparametersetpos_\primitiveindex = \mathrm{const}},
\end{equation}
\begin{equation}
    \label{eq:qpos}
    \charge^\poslabel_\primitiveindex\parens{\coordinatevector} =
    \sum_{\jprim\ne\primitiveindex} \indicator{\coordinatevector\in\primitivearea_\jprim} -
    \sum_{\ipix}\chargeraster \delta\parens{\coordinatevector - \coordinatevector_\ipix},
\end{equation}
\begin{equation}
    \label{eq:qsize}
    \charge^\sizelabel_\primitiveindex\parens{\coordinatevector} =
    \sum_{\jprim} \indicator{\coordinatevector\in\primitivearea_\jprim} -
    \sum_{\ipix}\chargeraster \delta\parens{\coordinatevector - \coordinatevector_\ipix},
\end{equation}
where \(\indicator{\blank}\) is the Iverson bracket, and \(\delta\) is the delta-function.

Expressions  \eqref{eq:fullpotential}-\eqref{eq:qsize} provide the physics-based foundation for our optimization: at every step, we use the new form of the energy terms to obtain the gradients using automatic differentiation; the ``frozen'' parts of each term are updated after parameter update at every step.  In this initial form, the functional has a number of undesirable properties for our application; we make several modifications described in the next section. 

\subsection{Discretization and functional modifications}
\subsubsection{Discretization.}
While for simple primitives the integrals 
in  \eqref{eq:fullpotential}-\eqref{eq:qsize} can be computed explicitly, 
we simplify the problem by using discrete charges instead of continuous distributions. 

The expression \eqref{eq:fullpotential} becomes equation (8) from the submission.
\begin{equation}
    \label{eq:Ei_discr}
    \iint\limits_{\canvasarea} \charge\parens{\coordinatevector_1}
    \iint\limits_{\primitivearea_{\primitiveindex}}
    \drpotential{\norm{\coordinatevector_1 - \coordinatevector_2}} \dprimitivearea{2}\dprimitivearea{1}
    \,\longrightarrow\,
    \sum_\ipix \charge_\ipix
    \iint\limits_{\primitivearea_{\primitiveindex}}
    \drpotential{\norm{\coordinatevector - \coordinatevector_\ipix}} \dprimitivearea.
\end{equation}

Expressions \eqref{eq:qpos} and \eqref{eq:qsize} become 
\begin{equation}
    \iint\limits_{\canvasarea} \indicator{\coordinatevector\in\primitivearea_\primitiveindex} f\parens{\coordinatevector} \dprimitivearea
    \quad\longrightarrow\quad
    \sum_\ipix \charge_{\primitiveindex,\ipix} f\parens{\coordinatevector_\ipix},
\end{equation}
\begin{equation}
    \iint\limits_{\canvasarea} \sum_{\ipix}\chargeraster \delta\parens{\coordinatevector - \coordinatevector_\ipix} f\parens{\coordinatevector} \dprimitivearea
    \quad\longrightarrow\quad
    \sum_\ipix \chargeraster f\parens{\coordinatevector_\ipix},
\end{equation}
\begin{equation}
    \label{eq:qpos_discr}
    \charge^\poslabel_\primitiveindex\parens{\coordinatevector} \quad\longrightarrow\quad
    \charge^\poslabel_{\primitiveindex,\ipix} = 
    \sum_{\jprim\ne\primitiveindex} \charge_{\jprim,\ipix} - \chargeraster,
\end{equation}
\begin{equation}
    \label{eq:qsize_discr}
    \charge^\sizelabel_\primitiveindex\parens{\coordinatevector} \quad\longrightarrow\quad
    \charge^\sizelabel_{\primitiveindex,\ipix} = 
    \sum_\jprim \charge_{\jprim,\ipix} - \chargeraster,
\end{equation}
where \(\chargeraster\) is the coverage of the  \(\ipix^\text{th}\) raster image pixel,
and \(\charge_{\kprim,\ipix}\) is the coverage of the  \(\kprim^\text{th}\) primitive 
in  \(\ipix^\text{th}\) pixel.

\subsubsection{Charge saturation.}
\begin{figure}[h!]
    \centerline{\includegraphics[width=\linewidth]{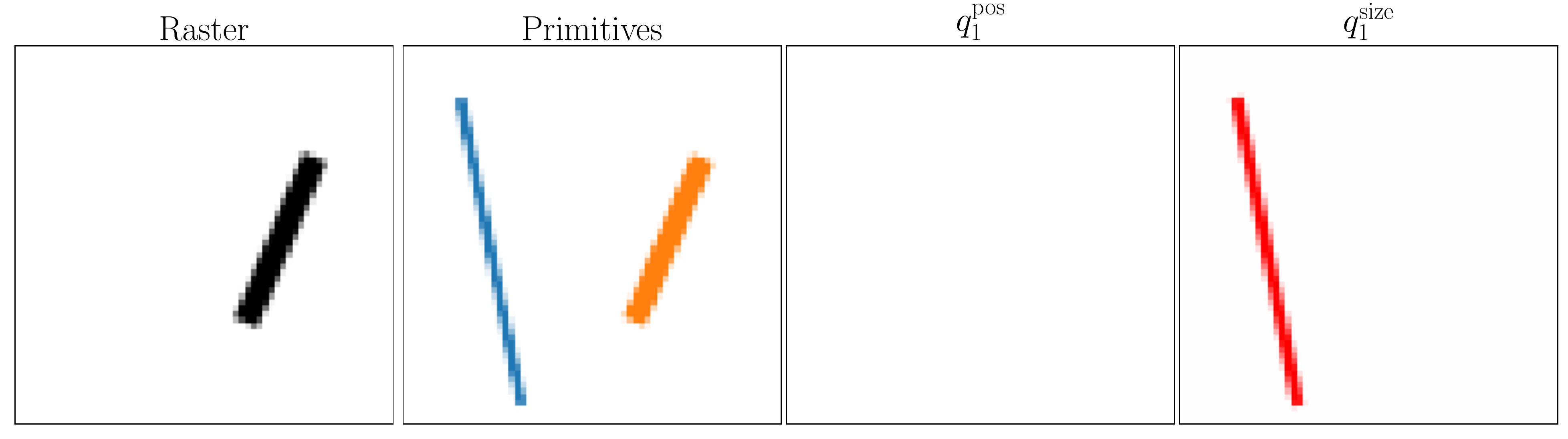}}
    \caption{Raster, primitives and charge grids of the first primitive. First primitive in blue, second in orange. In charge grids red represents excess charge.}
    \label{fig:sup_opt_ex_01}
\end{figure}
\begin{figure*}[h!]
    \centerline{\includegraphics[width=\linewidth]{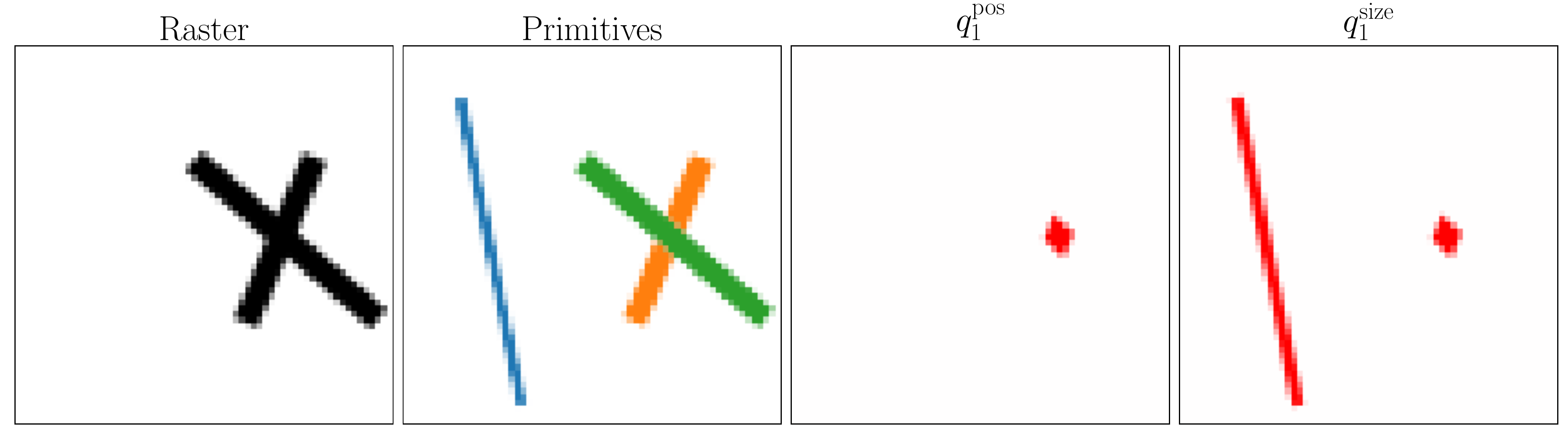}}
    \caption{Overlap affection on other primitives. First primitive in blue, second in orange. In charge grids red represents excess charge.}
    \label{fig:sup_opt_ex_02}
\end{figure*}
The charge distributions
\(\chargedistrib^\poslabel_\primitiveindex = \braces{\charge^\poslabel_{\primitiveindex,\ipix}}_\ipix\),
\(\chargedistrib^\sizelabel_\primitiveindex = \braces{\charge^\sizelabel_{\primitiveindex,\ipix}}_\ipix\)
are excess or insufficient charges that need to be compensated by 
changing  the \(\primitiveindex^\text{th}\) primitive. 
The energy terms corresponding to a primitive should not be affected by how many primitives cover a particular filled area.  For example, in  Figure~\ref{fig:sup_opt_ex_01}, the second primitive covers the filled part of the raster perfectly, and  this area does not affect the placement and size of the first primitive.   Figure~\ref{fig:sup_opt_ex_02}, the second and third primitives 
are covering the area equally well, but because of the overlap, the sum of their charges is higher than the negative charge of the raster image, and this creates a force acting on the first primitive.

To avoid the excess charge, we replace the sum of the charges with the maximum, leading to the following modification:  
\begin{equation}
    \label{eq:qpos_prefinal}
    \charge^\poslabel_{\primitiveindex,\ipix} = \chargetotalsub - \chargeraster,
\end{equation}
\begin{equation}
    \label{eq:qsize_final}
    \charge^\sizelabel_{\primitiveindex,\ipix} = \chargetotal - \chargeraster,
\end{equation}
where \(\chargetotal\) is the sum of coverages of \(\ipix^\text{th}\) pixel for all primitives, 
and \(\chargetotalsub\) is the same sum with  \(\primitiveindex^\text{th}\)  primitive excluded. 

Compared to \eqref{eq:qpos_discr}, \eqref{eq:qsize_discr}, modified charge distributions \eqref{eq:qpos_prefinal}, \eqref{eq:qsize_final} do not penalize overlaps. 

\newcommand\dinkus{\par\bigskip\noindent\hfill *\quad*\quad* \hfill\null\par\bigskip}

\paragraph{Illustrative examples.}
Next, we consider several examples illustrating the behavior of the functional, which also help us to explain the modifications we make. 

\noindent\emph{An isolated primitive.}
\label{sec:sup_opt_ex1}

\begin{figure}[h!]
    \centerline{\includegraphics[width=\linewidth]{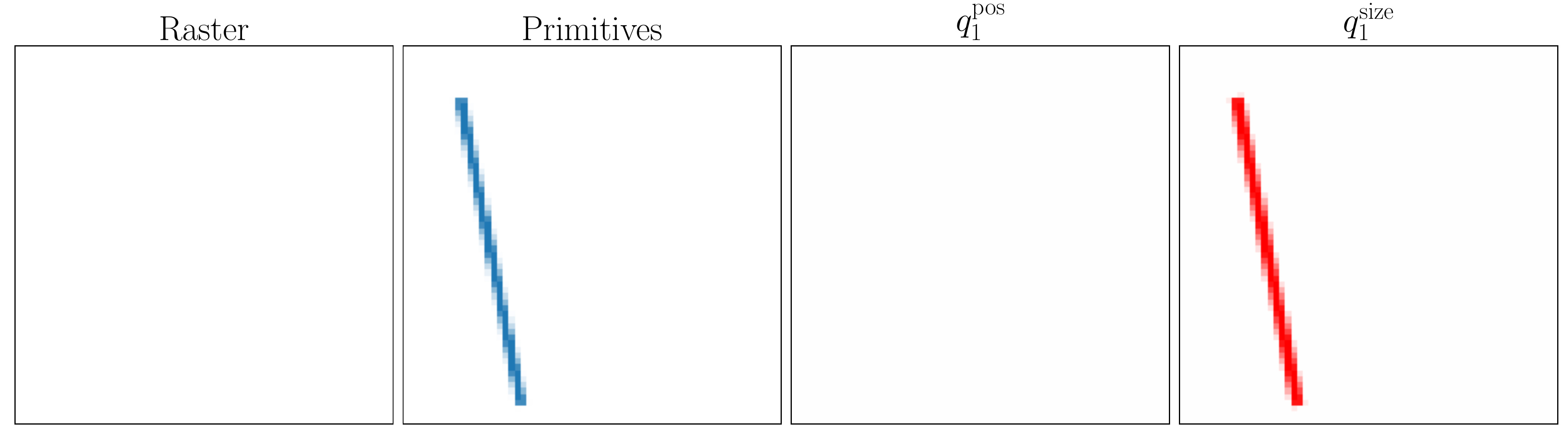}}
    \caption{Single primitive collapse. In charge grids red represents excess charge.}
    \label{fig:sup_opt_ex_1}
\end{figure}
For a single primitive (Figure~ \ref{fig:sup_opt_ex_1} ) the position energy term is constant and does not depend on the parameters, so it would not move. The size energy term becomes lower with size: the primitive collapses to a point.

\noindent\emph{Primitive separated from the filled areas of the raster image.}
\label{sec:sup_opt_ex2}
\begin{figure}[h!]
    \centerline{\includegraphics[width=\linewidth]{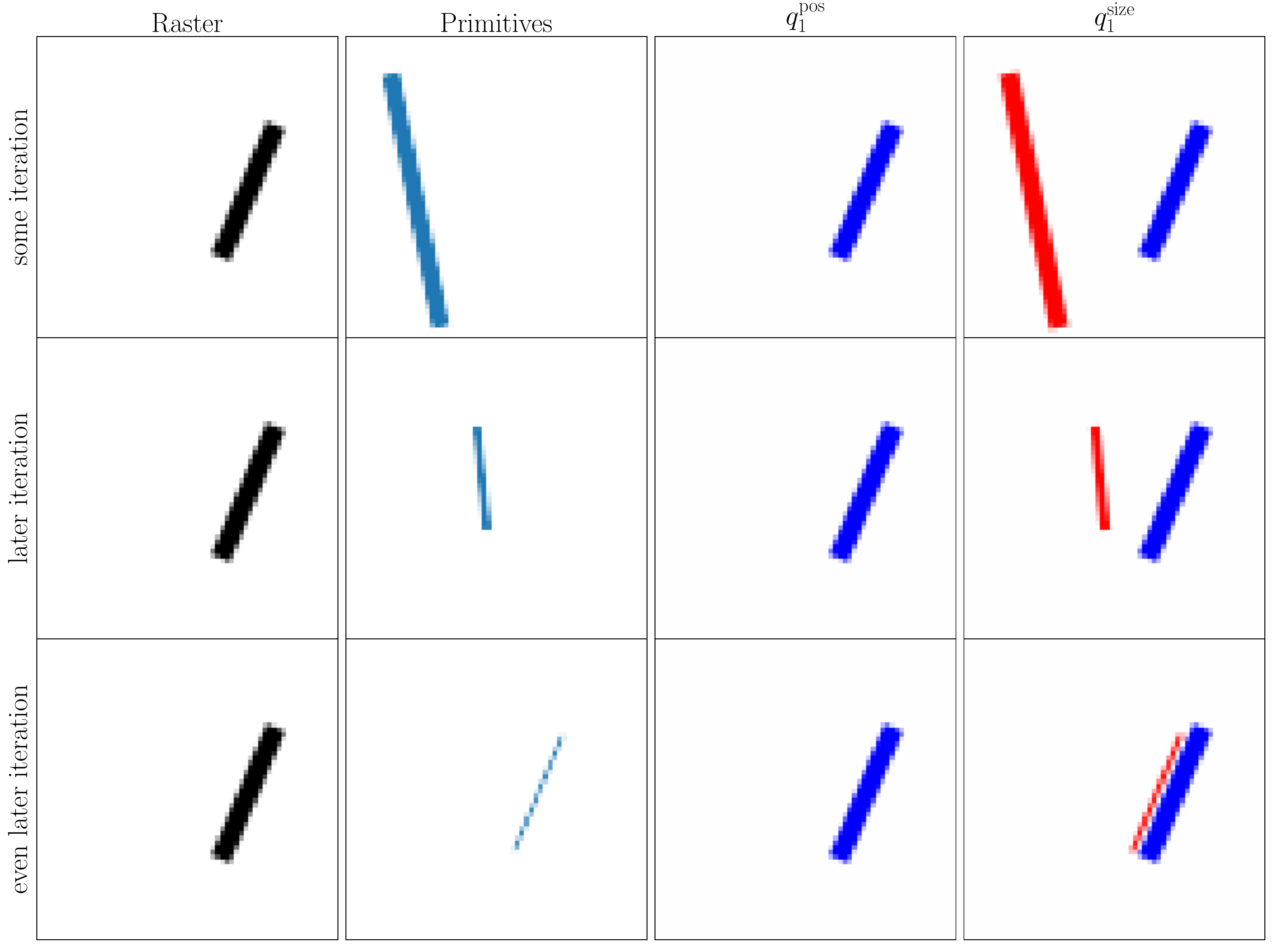}}
    \caption{Primitive interaction with far raster filled part. In charge grids red represents excess charge and blue represents uncovered raster.}
    \label{fig:sup_opt_ex_2}
\end{figure}
For a single primitive sufficiently far from the filled part of the image  (Figure~\ref{fig:sup_opt_ex_2}) the energy is decreasing if the primitive moves to the raster due to the second term in   \eqref{eq:qpos_prefinal}. If the primitive shrinks, the first term in \eqref{eq:qsize_final} decreases and the second term increases.  However, as we use fast-decaying potentials, we can neglect the interactions with distant charges, and overall the energy favors size reduction. 

\noindent\emph{A primitive close to a raster image.}
\label{sec:sup_opt_ex3}
\begin{figure}[h!]
    \centerline{\includegraphics[width=\linewidth]{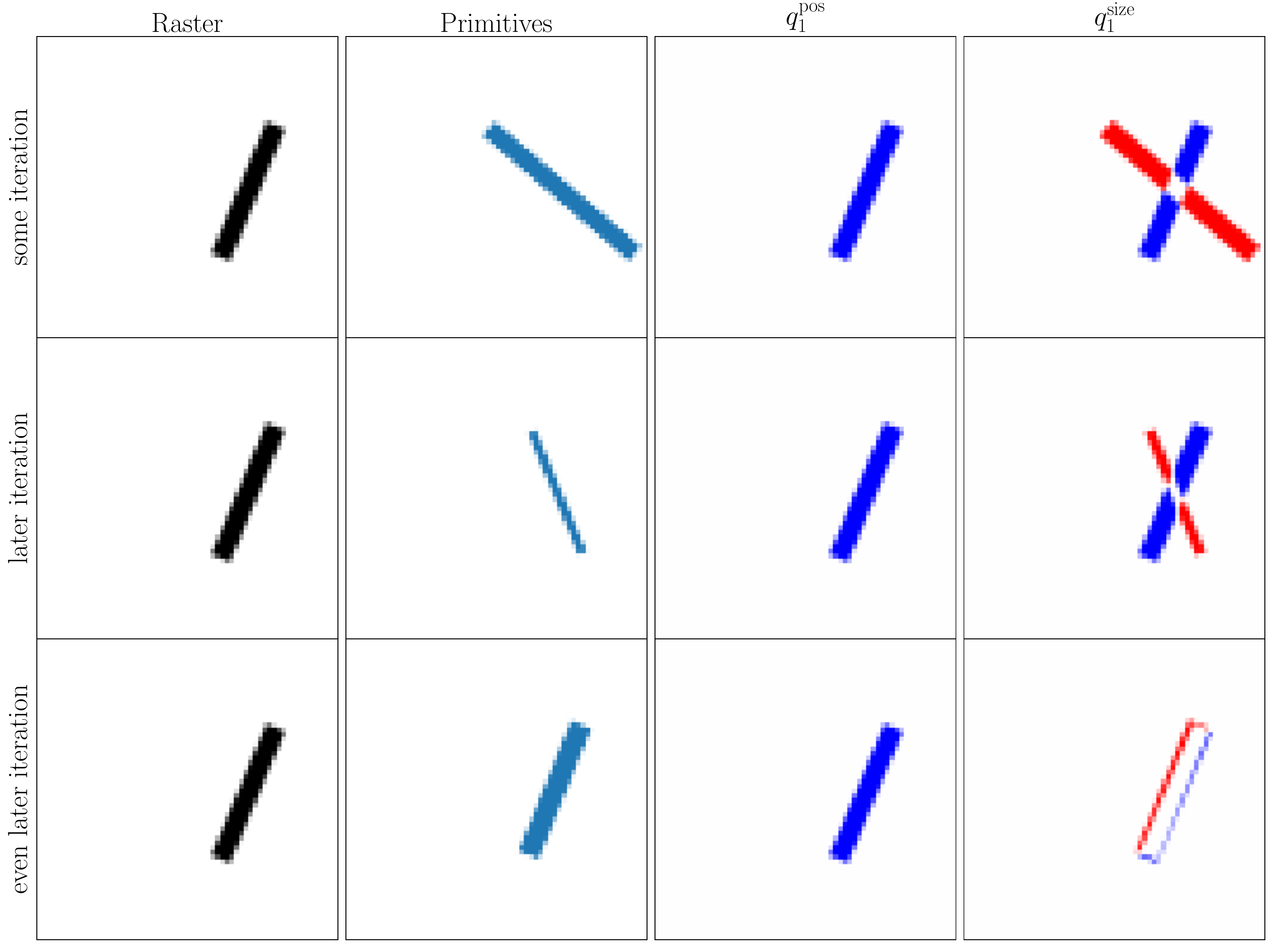}}
    \caption{Primitive interaction with close raster filled part. In charge grids red represents excess charge and blue represents uncovered raster.}
    \label{fig:sup_opt_ex_3}
\end{figure}
If a primitive is close to a filled part of the raster it will get aligned to the filled pixels, if these form a line and will increase in size until it covers the filled area (Figure~\ref{fig:sup_opt_ex_3})
If we add additional filled pixels or other primitives at a distance, due to potential decay, they will have a minimal effect on the behavior.

\noindent\emph{Primitive aligned with a filled part of the primitive.}
\label{sec:sup_opt_ex5}
In this case, the potentials from the raster image and the first primitive compensate each other, and the second primitive will not be affected by either, as in scenario of \ref{sec:sup_opt_ex1}.

\subsubsection{Enabling primitive intersections.}
\begin{figure*}[h!]
    \centerline{\includegraphics[width=\linewidth]{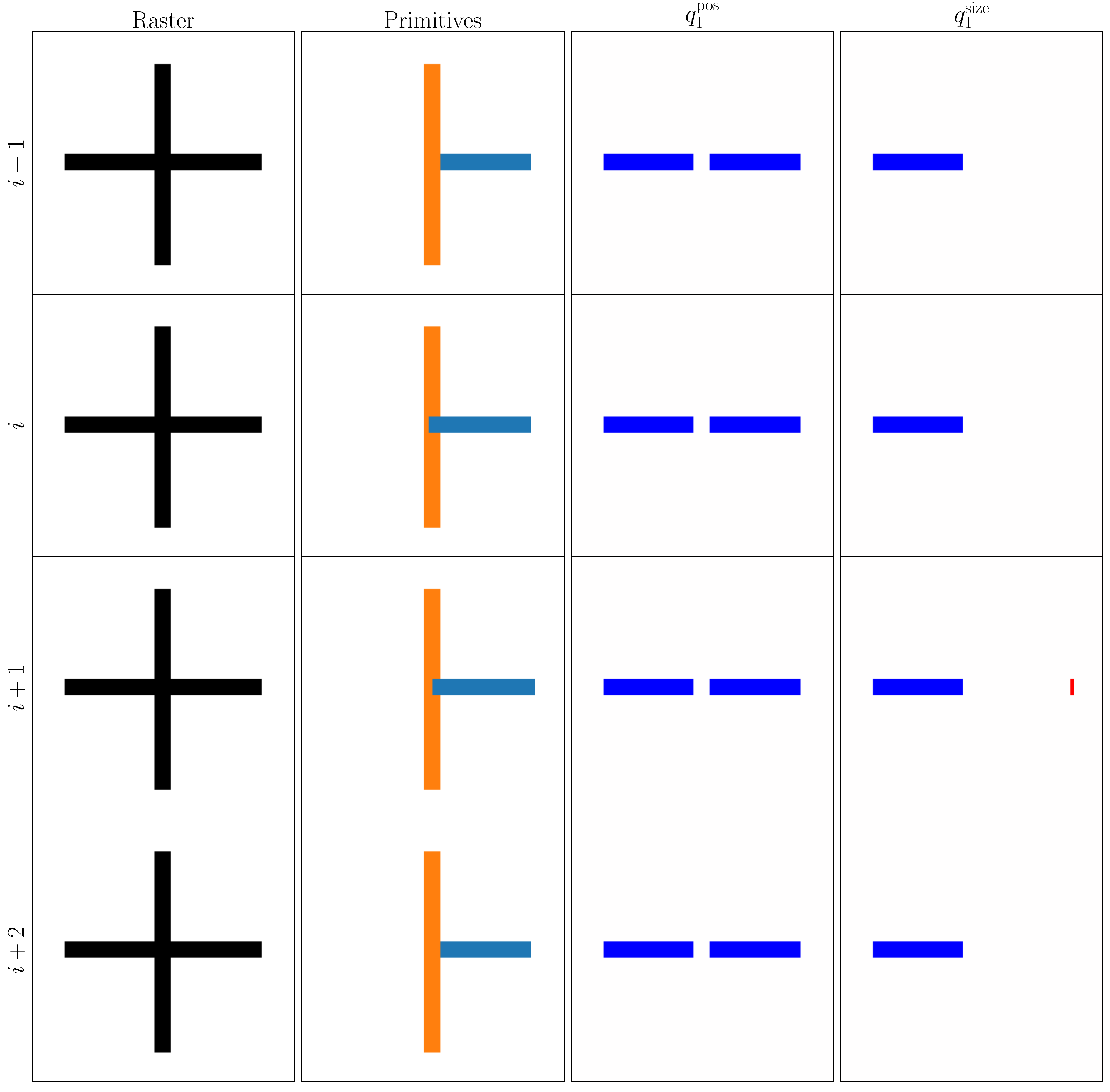}}
    \caption{Primitives interaction with disable primitive intersections. In charge grids red represents excess charge and blue represents uncovered raster.}
    \label{fig:sup_opt_ex_6}
\end{figure*}
Consider the case in  Figure~\ref{fig:sup_opt_ex_6}. In this case,it would be desirable for the second primitive to cover the entire horizontal line, but this will not happen, as the second primitive, 
with the original energy formulation will remain close to its initial state. 
Primitive 1 instead of expanding will not change much either, as primitive 2 prevents its expansion.

\begin{figure*}[h!]
    \centerline{\includegraphics[width=\linewidth]{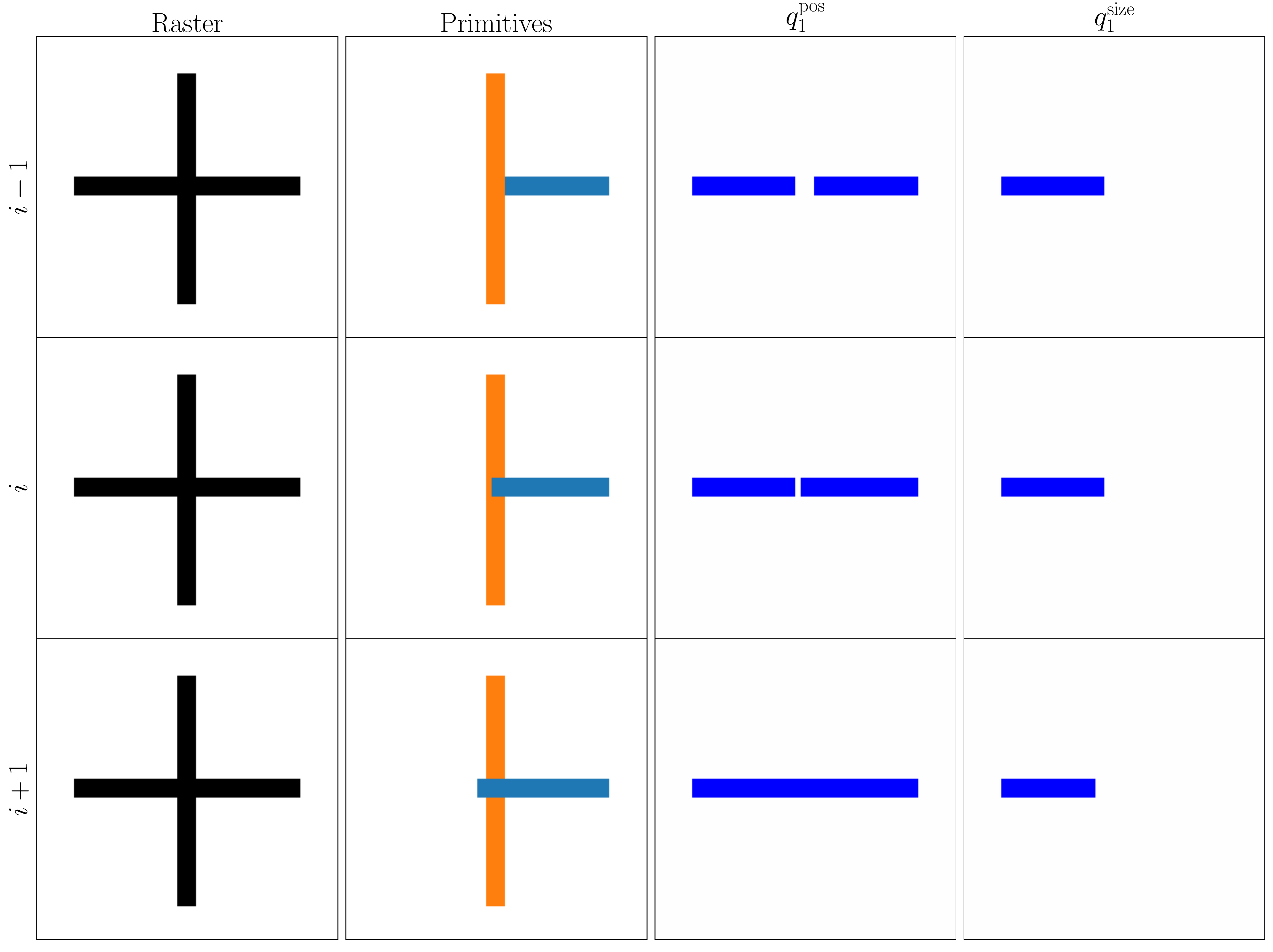}}
    \caption{Primitives interaction with enabled primitive intersections. In charge grids red represents excess charge and blue represents uncovered raster.}
    \label{fig:sup_opt_ex_61}
\end{figure*}

To achieve the desired effect, i.e., expansion of the second primitive, we modify \(\chargedistrib^\poslabel_\primitiveindex\) as follows:
\begin{equation}
    \label{eq:qpos_final}
    \charge^\poslabel_{\primitiveindex,\ipix} = \chargetotal - \charge_{\primitiveindex,\ipix} - \chargeraster.
\end{equation}
With this definition, the primitive interacts with pixels covered by other primitives, which allows it to expand across already filled areas, as in  Figure~\ref{fig:sup_opt_ex_61}

\subsubsection{Penalty for overlapping collinear primitives \(\edrredun\).}
\begin{figure*}[h!]
    \centerline{\includegraphics[width=\linewidth]{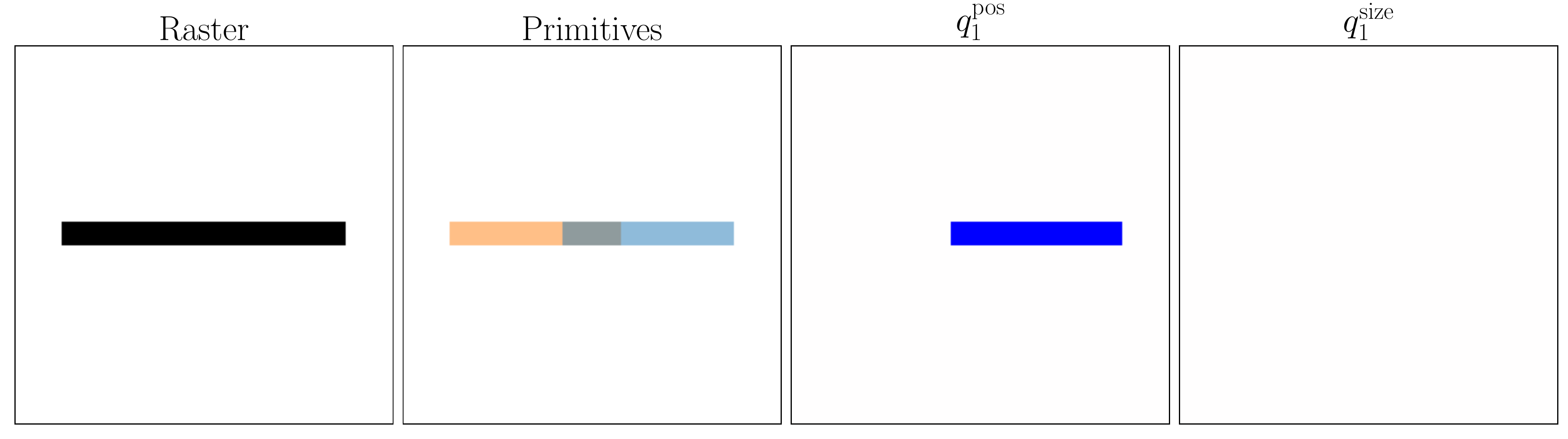}}
    \caption{Primitives covering the same area.}
    \label{fig:sup_opt_ex_7}
\end{figure*}
By itself, \eqref{eq:qpos_final} allows not just transversal intersections, but also aligned primitives covering the same area (Figure~\ref{fig:sup_opt_ex_7}). We add an additional penalty \(\edrredun\) to avoid this.  This term is also based on the interaction of the primitive 
with a background charge excess/deficiency, in this case, created by nearby primitives with tangents close to its tangent at close points.  We define this term for segments first, and then generalize to curves. 

\paragraph{Collinear penalty for line segments.} For a system of two segments
 \(\primitiveindex\) and \(\secondprimitiveindex\) we define it as
\def\alphacol{\alpha_\mathrm{col}}
\begin{equation}
    \edrredun = \withcondition{\edrpointprim\parens{\chargedistribredun}}{
        \text{close }\drpotential{\coordinatevectorlen},
        \primitiveparametersetpos_\primitiveindex = \mathrm{const}
    },
\end{equation}
\begin{equation}
    \chargeredunpix = \charge_{\secondprimitiveindex,\ipix}
    \exp{\parens{-\frac{
        \parens{\abs{\primitivedirection_\primitiveindex\cdot\primitivedirection_\secondprimitiveindex} - 1}^2}{
        \parens{\abs{\cos \alphacol} - 1}^2
    }}},
\end{equation}
where \(\text{close }\drpotential{\coordinatevectorlen}\) means that we truncate the interaction at 
a fixed radius  \(\drpotential{\coordinatevectorlen|\coordinatevectorlen > \coordinatevectorlen^*} = 0\), as explained below,  and charges \(\chargeredunpix\) are defined by  \(\secondprimitiveindex^\text{th}\)  primitive weighted by the cosine of the angle between segment directions,  \(\primitivedirection_\primitiveindex,\primitivedirection_\secondprimitiveindex\),
and \(\alphacol\) is a threshold angle for collinearity detection. 

For many segments, we define the charges as 
\begin{equation}
    \chargeredunpix = \norm{\excessvectorfieldpix}
    \exp{\parens{-\frac{
        \parens{\abs{\primitivedirection_\primitiveindex\cdot\excessvectorfieldpix} - 1}^2}{
        \parens{\abs{\cos \alphacol} - 1}^2
    }}},
\end{equation}
where \(\excessvectorfieldpix = \sum_{\secondprimitiveindex\ne\primitiveindex} \primitivedirection_\secondprimitiveindex \charge_{\secondprimitiveindex,\ipix} \)
is the sum of directions of all the other primitives weighted \wrt the mean direction of other primitives.

\paragraph{Collinearity penalization for curved segments.}
For curves, we use a similar idea, but need to use a different direction for every pixel, 
 \(\primitivedirection_{\primitiveindex,\pixelindex}\) 
\begin{equation}
\begin{gathered}
    \chargeredunpix = \norm{\excessvectorfieldpix}
    \exp{\parens{-\frac{
        \parens{\abs{\primitivedirection_{\primitiveindex,\ipix}\cdot\excessvectorfieldpix} - 1}^2}{
        \parens{\abs{\cos \alphacol} - 1}^2
    }}},\\
    \excessvectorfieldpix = \sum_{\secondprimitiveindex\ne\primitiveindex} \primitivedirection_{\primitiveindex,\ipix} \charge_{\secondprimitiveindex,\ipix}.
\end{gathered}
\end{equation}
This definition reduces to the definition for segments if the curve is straight. 

\subsubsection{Connected area mask.}
\begin{figure*}[h!]
    \centerline{\includegraphics[width=\linewidth]{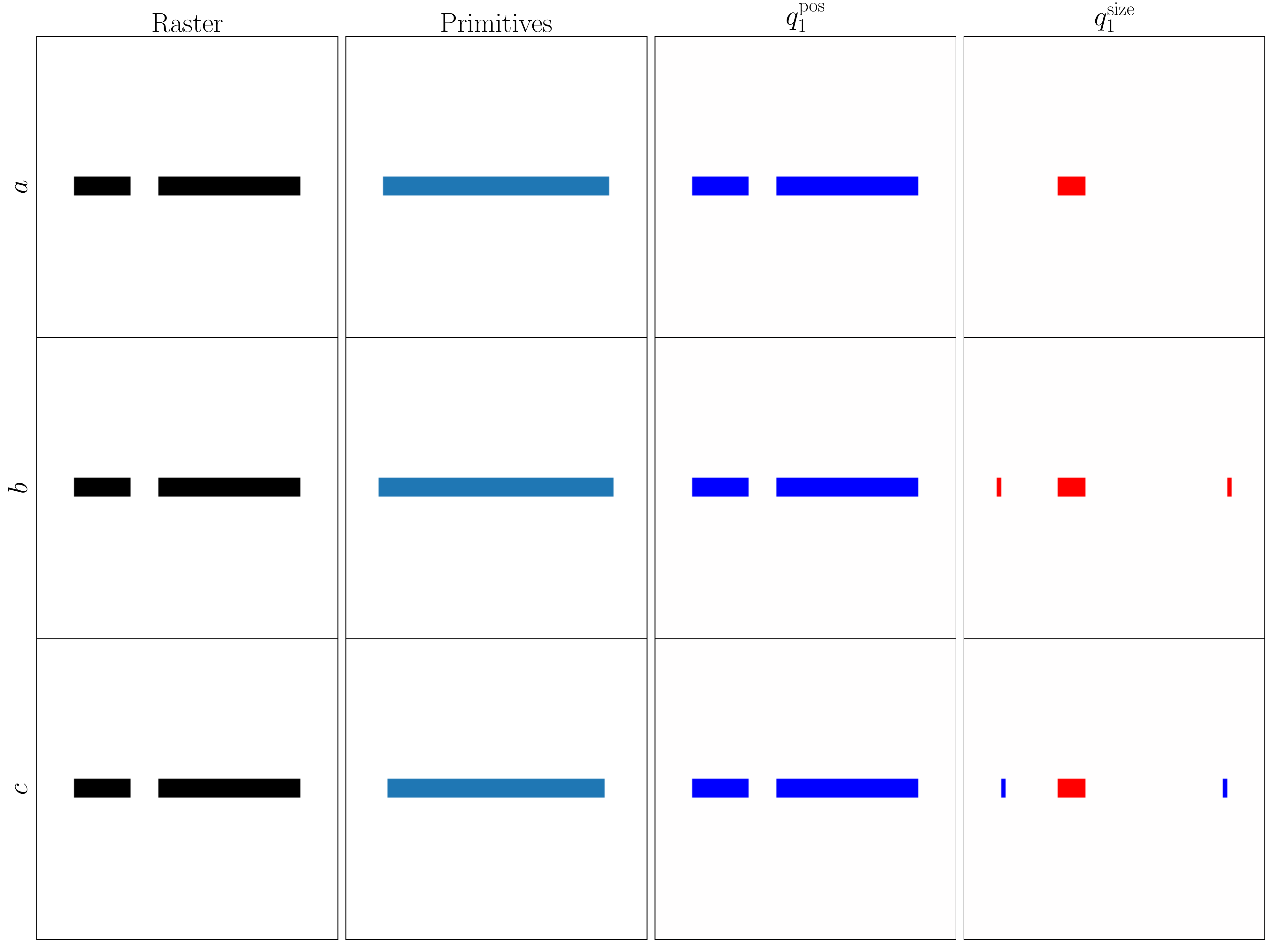}}
    \caption{Undesirable local minimum in which a primitive covers two disconnected raster parts.}
    \label{fig:sup_opt_ex_H}
\end{figure*}
\begin{figure*}[h!]
    \centerline{\includegraphics[width=.3\linewidth]{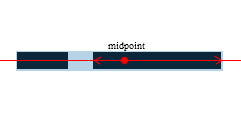}\quad%
    \includegraphics[width=.3\linewidth]{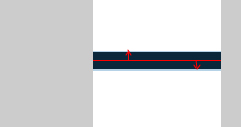}\quad%
    \includegraphics[width=.3\linewidth]{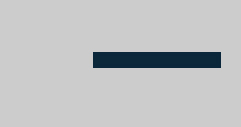}}
    \caption{Stages of \(\edrcoef_{\primitiveindex,\ipix}\) calculation.}
    \label{fig:sup_opt_ck_calc}
\end{figure*}

Consider the example in Figure~\ref{fig:sup_opt_ex_H}~(a). Clearly the position and width of the primitive cannot be changed so that the energy decreases.  Same is true for length:
If the length increases or decreases (Figure~\ref{fig:sup_opt_ex_H}~(b,c)), a counteracting force immediately appears because of the charge excess or deficit. We conclude that this is a local minimum,
but clearly not a desirable solution. To reduce the chances of a primitive getting stuck in such a minimum, we limit the interaction of the primitive with the raster to the part that it can cover for its current position and orientation, but arbitrary size. To be more precise, for line segments
 (Figure~\ref{fig:sup_opt_ck_calc}), for a fixed position and orientation, we find unfilled pixels 
 along the line of the segment closest to its center, determining the length of the area. Once this is determined, then we find  unfilled pixels closest to the line of the segment on two sides. This determines the rectangle for which  the mask coefficient \(\edrcoef_{\primitiveindex,\ipix}\) is set to one, and for the rest of the image to zero.  

For second order B\'ezier curves, we use a similar definition. Instead of a line we use a parabola containing the B\'ezier segment, and the distance along the parabola instead of the Euclidean distance.

The charge distribution for the size terms of the energy is  redefined as follows
\begin{equation}
    \label{eq:qsize_c}
    \charge^\sizelabel_{\primitiveindex,\ipix} =
    \begin{cases}
        \chargetotal - \chargeraster     &  \text{if } \edrcoef_{\primitiveindex,\ipix} = 1, \\
        \charge_{\primitiveindex,\ipix}  &  \text{if } \edrcoef_{\primitiveindex,\ipix} = 0.
    \end{cases}
\end{equation}

\subsubsection{Connected area mask for positional terms.}
If we use the same masks for the charges used for the positional terms
\(\charge^\poslabel_{\primitiveindex,\ipix}\) eliminating the influence of everything 
outside the area where   \(\edrcoef_{\primitiveindex,\ipix}\) is positive, then the primitive 
will stay within the filled area they initially overlap, which may be undesirable if the initial 
position is inaccurate, or multiple primitives initially cluster in the same place.
For this reason, we only amplify the charge in the masked area leaving it the same outside:

\def\ckposcoef{\lambda_\mathrm{pos}}
\begin{equation}
    \label{eq:qpos_c}
    \charge^\poslabel_{\primitiveindex,\ipix} =
    \begin{cases}
        \ckposcoef\parens{\chargetotal - \charge_{\primitiveindex,\ipix} - \chargeraster}     &  \text{if } \edrcoef_{\primitiveindex,\ipix} = 1, \\
        \chargetotal - \charge_{\primitiveindex,\ipix} - \chargeraster  &  \text{if } \edrcoef_{\primitiveindex,\ipix} = 0,
    \end{cases}
\end{equation}
The amplification coefficient \(\ckposcoef\) is chosen empirically.

\subsubsection{The choice of the potential function.}
The main property of the potential function \(\drpotential{\coordinatevectorlen}\)
used in our algorithm is rapid monotonic decrease with distance.
We use the function
\begin{equation}
    \label{eq:sup_pointpotential}
    \drpotential{\coordinatevectorlen} = \
            e^{-\frac{\coordinatevectorlen^2}{\rclose^2}}\
          + \cfar e^{-\frac{\coordinatevectorlen^2}{\rfar^2}},
\end{equation}
which allows us to control interactions at close range \(\sim\rclose\)
independently from interactions at far range \(\sim\rfar\).
We choose these ranges and the weight experimentally
\(\rclose = 1\,\mathrm{px}\), \(\rfar = 32\,\mathrm{px}\), \(\cfar = 0.02\),
and in \(\edrredun\) disable the far interactions by setting \(\cfar=0\).
\end{document}